\documentclass[final,5p,times,twocolumn]{elsarticle}

\usepackage{amssymb}
\usepackage{amsmath}
\usepackage{graphicx}
\usepackage{caption}
\usepackage{subcaption}

\usepackage{algorithm}
\usepackage{algpseudocode}

\usepackage{hyperref}
\usepackage[none]{hyphenat} 

\makeatletter
\def\ps@pprintTitle{%
  \let\@oddhead\@empty
  \let\@evenhead\@empty
  \def\@oddfoot{%
    \footnotesize 
    \parbox[t]{\textwidth}{%
      \copyright~2025. This manuscript version is made available under the CC-BY-NC-ND 4.0 license. 
      The final version of record is available at \href{https://doi.org/10.1016/j.robot.2025.105059}{https://doi.org/10.1016/j.robot.2025.105059}.
    }
  }%
  \let\@evenfoot\@oddfoot
}
\makeatother

\begin{document}

\begin{frontmatter}

\title{Optimizing Exploration with a New Uncertainty Framework for Active SLAM Systems}

\author[add1]{Sebastian Sansoni}
\ead{ssansoni@inaut.unsj.edu.ar}
\author[add1]{Javier Gimenez}
\ead{jgimenez@inaut.unsj.edu.ar}
\author[add2]{Gastón Castro}
\ead{gcastro@cifasis-conicet.gov.ar}
\author[add1]{Santiago Tosetti}
\ead{stosetti@inaut.unsj.edu.ar}
\author[add1]{Flavio Capraro}
\ead{fcapraro@inaut.unsj.edu.ar}

\author{}
\address[add1]{Instituto de Automática (INAUT), Universidad Nacional de San Juan - CONICET, J5400ARL, San Juan, Argentina}
\address[add2]{CIFASIS, French Argentine International Center for Information and Systems Sciences (CONICET $-$ UNR), Rosario, Argentina}

\begin{abstract}	
    Accurate reconstruction of the environment is a central goal of Simultaneous Localization and Mapping (SLAM) systems. However, the agent’s trajectory can significantly affect estimation accuracy. 
    This paper presents a new method to model map uncertainty in Active SLAM systems using an Uncertainty Map (UM). The UM uses probability distributions to capture where the map is uncertain, allowing Uncertainty Frontiers (UF) to be defined as key exploration-exploitation objectives and potential stopping criteria. In addition, the method introduces the Signed Relative Entropy (SiREn), based on the Kullback–Leibler divergence, to measure both coverage and uncertainty together. This helps balance exploration and exploitation through an easy-to-understand parameter.

    Unlike methods that depend on particular SLAM setups, the proposed approach is compatible with different types of sensors, such as cameras, LiDARs, and multi-sensor fusion. It also addresses common problems in exploration planning and stopping conditions.  Furthermore, integrating this map modeling approach with a  UF-based planning system enables the agent to autonomously explore open spaces, a behavior not previously observed in the  Active SLAM literature.

    Code and implementation details are available as a ROS node, and all generated data are openly available for public use, facilitating broader adoption and validation of the proposed approach.
\end{abstract}


\end{frontmatter}

\section{Introduction}\label{sec:intro}
Active SLAM (\textit{Simultaneous Localization and Mapping}) or ASLAM involves generating trajectories for an agent, such as an autonomous robot, or a teleoperated robot, intending to gather data to produce a high-quality map and precise agent localization. The definition of \textit{quality} varies depending on different approaches. For instance, in some scenarios, quality may be quantified based on how extensively and rapidly an area is covered by the sensor. In other cases, conserving energy may be considered the most critical factor. In this work, we define quality in terms of the uncertainty of the map. 

It is well known that the trajectory followed by the sensor significantly impacts the precision of the states estimated by the SLAM system~\cite{khosoussi2014novel}. If this trajectory is not carefully planned, the uncertainty of the map becomes unpredictable. In this context, planning a path forward involves addressing the exploration-exploitation dilemma, which reflects the trade-off between exploiting known information and exploring unknown sites to incorporate new data. In unknown environments, exploration is prioritized, raising uncertainty until previously mapped areas are revisited, where a loop closure may occur.

Since Yamauchi's work~\cite{yamauchi1997frontier}, frontiers, defined as the region separating the explored from the unexplored, have been widely used as exploration objectives. In~\cite{stachniss2005information}, the concept of map and trajectory entropy is included, establishing a balance between exploring and exploiting. Aligned with this idea of entropy,~\cite{leung2006active,atanasov2015decentralized} proposes incorporating artificial points as attractor regions to encourage the trajectory planning system to visit or explore specific sites on the map. These points are generally characterized by their high uncertainty, which causes elevated local entropy. 

Additionally, since path planning requires knowledge of the location and shape of obstacles, the use of occupancy maps for spatial representation has played a crucial role in both historical and contemporary ASLAM strategies in 2D~\cite{elfes1989using, thrun2003learning, stachniss2005information} and, less commonly, in 3D using systems like OctoMap~\citep{hornung2013octomap}. Incorporating complementary information into occupancy maps enables trajectory optimization based on the quality of data obtained at each cell~\cite{moorehead2001autonomous, bourgault2002information, stachniss2003exploring, o2012gaussian}. In line with this philosophy, this work proposes using an occupancy map for the planning system and, in addition, introduces a new map, called the \textit{Uncertainty Map} (UM), to calculate a quality index and identify new exploration objectives. 

The quality measure, inspired by the \textit{Kullback-Leibler divergence} $(D_{KL})$, quantifies the information stored in the uncertainty map relative to a reference map. This measure, called \textit{Signed Relative Entropy} (SiREn), is used to evaluate map's quality and coverage. 

Moreover, the UM is used to define the concept of \textit{Uncertainty Frontiers} (UF), which are map regions characterized by significant local discontinuities in exploration quality. These frontiers serve as generators of exploration objectives and establish stopping criteria for ASLAM systems. Exploration targeted at UF aims to smooth these discontinuities. Furthermore, the UF concept helps identify well-known areas that need revisiting to reduce the map’s uncertainty. Additionally, the limit to which the UM can be smoothed may serve as a stopping criterion, which is significant given that it remains an open problem~\cite{placed2022enough, placed2023survey}. 

Interpretability is a crucial property of a model or system, particularly in decision making processes, alowing the user understanding the relationships among variables. This attribute is essential because a lack of interpretability can lead to misunderstandings about how decisions are made, raising concerns over the accuracy and reliability of the results. Therefore, ensuring interpretability is vital for making models and systems effective and trustworthy in decision-making contexts~\cite{aguirre2023, nauck2003measuring}.  The exploration-exploitation trade-off typically involves employing dimensionless weighting factors adjusted heuristically~\cite{carrillo2018autonomous}. In contrast, the methodology proposed in this work only requires a readily interpretable and measurable weighting parameter. 

In summary, this work proposes:
\begin{itemize}
    \item A novel method for modeling map uncertainty.
    \item A quality index for the map based on uncertainty levels.
    \item A new generator of exploration-exploitation objectives that can serve as stopping criteria in ASLAM systems.
\end{itemize}
The source code is implemented in the \textit{Robot Operating System} (ROS) framework and is publicly available\footnote{\href{https://github.com/Seba-san/UncertaintyMap}{https://github.com/Seba-san/UncertaintyMap}}.

This work is organized as follows. In Sec.~\ref{sec:related} the state of the art regarding the uncertainty modeling with occupancy maps is discussed.
Uncertainties are characterized through probabilities in Sec.~\ref{sec:exploration_map}, and the updating process for these probabilities is detailed in Sec.~\ref{sec:log-odds}.
The procedure for obtaining the uncertainty map from the probability map, along with the concept of uncertainty frontiers as a generator of exploration objectives, is described in Sec.\ref{sec:uncertainty_frontiers}.
The method for obtaining the signed relative entropy index is described in Sec.~\ref{sec:relative_entropy}.
To complete the theoretical development, a computational cost analysis is carried out in Sec.\ref{sec:computational_cost}.
A series of experiments are developed in Sec.~\ref{sec:experiments} to validate different aspects of this proposal.
Finally, conclusions are presented in Sec.~\ref{sec:conclusions} and implementation details are included in the Appendices.


\section{Related Works}\label{sec:related}

This work introduces a toolkit designed to help active SLAM systems. It includes a new method for modeling uncertainty, a metric to assess this feature, and the identification of new exploration objectives. The presented literature review addresses widely used strategies and seeks to identify structural weaknesses in these approaches that affect the generalization of results, replicability, and the ability to compare different techniques. These issues are identified as general problems in ASLAM systems~\cite{placed2023survey}. 

In~\cite{carrillo2012comparison}, an analysis is conducted on various criteria for measuring the degree of uncertainty, whether of the agent, the trajectory, or the map. Through the work of Kiefer~\cite{kiefer1974general}, it is known that the A-\textit{opt}, D-\textit{opt}, and E-\textit{opt} criteria have a common origin, and thus, they share certain properties. Among these, D-\textit{opt} is the only one that is proportional to the volume of the uncertainty ellipse of the estimated variables, and remains invariant under reparameterization and linear transformations. Moreover, in the context of absolute uncertainty representation, D-\textit{opt} is the only criterion that increases monotonically as overall uncertainty grows. This property, demonstrated in~\cite{carrillo2015monotonicity,rodriguez2018importance}, ensures that the measure reliably captures variations in uncertainty. As a result, it is particularly useful for ASLAM applications.

Subsequently the normalized D-\textit{opt} criterion for a covariance matrix $\mathbf{\Sigma} \in M^{n\times n}$ is: 
\begin{align}
    D\text{-\textit{opt}}(\mathbf{\Sigma})&=|\mathbf{\Sigma}|^\frac{1}{n}=\text{exp}\Bigg(\frac{1}{n} \sum_{i=1}^{n} \log (\lambda_i)  \Bigg),
\end{align}
where $\lambda_i$ is the $i$-th eigenvalue of $\mathbf{\Sigma}$ and $|\cdot|$ is the determinant operator. This criterion offers a geometric interpretation, understood as the $n$-th root of the volume of the hyperellipsoid associated with the covariance matrix. This approach, however, is only valid when the variables under analysis are in length units, representing the semi-axes of the hyperellipsoid. For other units, such as angular amplitudes, the model’s interpretability is lost, making it difficult to explain the underlying phenomena.
Additionally, without a specific reference value, the utility of the $D\text{-\textit{opt}}(\mathbf{\Sigma})$ measure, which ranges from $(0,\infty)$, remains ambiguous. Moreover, while the covariance matrix defines the fit of an $n$-order hyperboloid to the data, this measure may not be suitable if the data are not Gaussian. In contrast, this work proposes a scale-invariant uncertainty measure with values between $0$ and $1$, retaining interpretability and generality. This improvement makes the results more transparent and reliable for the user, and it integrates non-Gaussian measurements, broadening the applicability of this criterion. 

Similarly, Shannon entropy is one of the most widely used utility functions for grid maps today~\cite{ahmed2023active}. Its use was first proposed in~\cite{stachniss2005information} and later improved in~\cite{stachniss2006exploration}. It is defined as: 
\begin{align}\label{eq:shannon}
    H(\mathbf{m})=-r^2 \sum_{c\in \mathbf{m}}p(c) \log p(c) + \Big(1-p(c)\Big) \log \Big(1-p(c)\Big),
\end{align}
where $H(\mathbf{m})$  represents the Shannon entropy of the map $\mathbf{m}$, $p(c)$ is the Bernoulli occupation probability for the cell $c$ and $r^2$ denotes the area covered by each cell. 
An alternative to Shannon entropy is proposed in~\cite{blanco2008novel} to address specific implementation challenges, such as the significant imbalance where the map’s entropy exceeds the agent’s or trajectory’s entropy by more than an order of magnitude. 
These dimensional differences, along with the essential use of numerically sensitive heuristic parameters to address them, are reported in~\cite{carrillo2018autonomous}. 
Additionally, Shannon entropy is not well-definded for continuous functions, leading to compatibility issues when combining the entropies of the agent’s pose and the map. 
Even for the same trajectory, entropy values can vary depending on the map's discretization (by adjusting $r$ in (\ref{eq:shannon})) and the number of unexplored cells. Essentially, entropy escalates as more areas remain uncovered. These fluctuations stem from the map modeling rather than any alterations in the information held by the system. 
This complicates comparisons between different methods or different stages of the same method when there are variations in discretization (e.g., with OctoMap~\cite{hornung2013octomap}) or in the number of unexplored cells. Due to this ambiguity in the results, its generalization to ASLAM becomes challenging~\cite{placed2023survey}. 

To address the issue of the map's entropy significantly exceeding that of the agent or trajectory,~\cite{carrillo2018autonomous} proposes a utility function based on Rényi entropy, defined as follows: 
\begin{align}\label{eq:renyi}
    H_{\alpha}(\mathbf{m})=\frac{1}{1-\alpha}\log_2\left(\sum_{c\in \mathbf{m}}p(c)^{\alpha}\right),
\end{align}
where $\alpha$ is a parameter that defines the order of the entropy. A utility function for exploration is then proposed as: 
\begin{align}
I_{c}(\mathbf{m})&=H_{\alpha=1}(\mathbf{m})-H_{\alpha=c}(\mathbf{m}),
\end{align}
where, $I_{c}(\mathbf{m})$ represents the mutual information, $H_{\alpha=1}(\mathbf{m})$ denotes the Shannon entropy of the map, and $H_{c}(\mathbf{m})$ indicates the Rényi entropy of the map with parameter $c$. This parameter acts as a weighting factor and relates to the  agent's confidence level in its trajectory. The proposed approach incorporates this factor within the entropy framework, enabling a selective balance in information gain.  Which weighting benefits actions that reduce the agent's uncertainty and nullify those not providing significant information. During transitions,  information gain decreases as the agent's uncertainty increases. This behavior resembles the one described in this work;  however, SiREn becomes negative when the agent's uncertainty exceeds the preferred maximum. Nevertheless, as detailed above, the proposal in \cite{carrillo2018autonomous} shares many of Shannon entropy’s limitations.

In this work, a unified theory for measure the \textit{signed relative entropy} of a map is proposed, which is immune to discretization changes and the number of unexplored cells. Additionally, this approach fuses the agent's uncertainty with the map's uncertainty through the sensor's field of view. Given that this method does not depend on the SLAM used, it provides a tool that allows for the generalization of results and comparison between different ASLAM methods or stages of the same system. Furthermore, it offers a measure that integrates the explored area with its measurement uncertainty,  an advantage over other methods that assess uncertainty and coverage separately. 

On the other hand, the following research studies propose advances in occupancy mapping methodologies, improving planning effectiveness and demonstrating that these maps alone are insufficient for comprehensive planning tasks. In \cite{moorehead2001autonomous}, a grid map is proposed where each cell contains two information vectors: one vector stores the sensor information related to that cell, and the other indicates whether the cell is navigable. Additionally, this approach assesses the need to explore an area by estimating the associated information gain, which requires weighing the contribution of each sensor. The proposed method includes several parameters that must be adjusted heuristically. Also \cite{bourgault2002information} utilizes an occupancy grid along with sparse feature SLAM and, given an action, considers the expected information gain from both. A linear combination weighs and fuses both information gains with predefined values that must be tuned heuristically for each scenario.  

A novel concept of coverage maps is introduced in \cite{stachniss2003exploring}. Unlike the classical binary occupancy representation, where cells are categorized as occupied or not, this work proposes assigning occupancy probability to each cell. Exploration continues until a cell's uncertainty falls below a specified threshold or remains insufficiently reduced after five measurements. 
The study concludes that the optimal balance between exploitation and exploration is achieved by minimizing both maximum uncertainty and traveled distance. In \cite{stachniss2005information}, map uncertainty is integrated with trajectory uncertainty. This study considers a binary occupancy map and estimates the amount of information as the sum of the Shannon entropy in each cell. The mean entropy of the map is used to predict the amount of information in unknown spaces. Trajectories are selected from a set of classical frontier locations and areas with high probabilities of loop closure. 
In~\cite{amigoni2010information}, a utility function is proposed by fusing the distance traveled and the predicted relative entropy change resulting from a specific action. This fusion is carried out without weighting factors, which makes the system improves its  robustness. The approach uses points within the field of view as exploration objectives, with the stopping criterion based on their accessibility and other factors.

The addition of an uncertainty map to the occupancy map was proposed in~\cite{o2012gaussian} and later integrated into ASLAM systems in~\cite{jadidi2015mutual, ghaffari2018gaussian}. This system employs Gaussian Processes to Model Occupancy Maps (\textit{GPOMs}). The method involves solving a regression problem for a binary classification of cells as occupied or non-occupied using a Gaussian kernel. This approach is implemented through a neural network, which attempts to estimate the hyperparameters of the kernels. These include the variance of the observations associated with a position,which proves valuable for planning tasks.
In a later work \citep{ghaffari2018gaussian}, the same concept of using Gaussian processes to generate an uncertainty map and a probabilistic frontier map is applied. However, this approach entails high computational costs depending on the number of measurements required to determine the Gaussian kernel's hyperparameters. Additionally, the adjustment is performed using a Gaussian kernel, which may not be the most effective method for all systems. In contrast, this work proposes a method to model the uncertainty map with lower computational costs and without assuming any specific distribution, improving the generality of the application. 

The final approach to mention is the concept of frontiers, proposed by Yamauchi \cite{yamauchi1997frontier}, which has been widely adopted by most planning systems either as exploration objectives or as an integral component of the entire system \cite{ahmed2023active}. To adapt this concept for optimal trajectory planning systems based on information, a component with high entropy is added to the frontiers, rewarding the planner for reaching them \cite{leung2006active, atanasov2015decentralized}. These attractor points or regions are generally characterized by high uncertainty, which leads to elevated local entropy and motivates the agent to revisit or explore these specific sites on the map. This modification to Yamauchi’s frontiers enables further changes, such as those proposed in this work. The definition of a new map that complements the occupancy maps allows the introduction of a novel frontier concept,  achieving results with lower variability and improved performance. 

Readers interested in a broader overview of ASLAM are directed to~\cite{placed2020deep, placed2023survey, ahmed2023active}, among other sources.

\section{Uncertainty Framework}\label{sec:theoretical_framework} 

Each measurement only approximates the true value, and it is considered complete only when its uncertainty has been estimated~\cite{iso1995guide}. In this context, an occupancy map represents the result of a measurement, while the UM serves as an estimate of the associated uncertainties. 

The uncertainty in a measurement is related to the dispersion of its distribution, which is associated to an specific probability known  as  dispersion probability. These probabilities are stored in an array called a probability map, and the procedure for building and updating it is explained below. Subsequently, the concepts of UF and UM are introduced. The SiREn is defined to compare the quality of different maps, landmarks, and trajectories. 

\subsection{Dispersion probability and uncertainty}
\label{sec:exploration_map}

Uncertainty is defined as the dispersion of a variable around its central measures, such as the mean, median, and mode.
When measurements are modeled with an unimodal random variable $X$, they are concentrated in a region around its mode/mean with probability inversely related to the measurement's dispersion. The inequalities originally proposed by Chebyshev and later refined by Gauss \cite{van2016generalized} provide a bound from this probability. Specifically, if $X$ is a random variable (RV) with variance $\sigma^2$ and mode equal to its mean $\mu$, then: 
\begin{align}\label{eq:gauss_desig}
P\left(|X-\mu|<\frac{s}{2}\right)\geq \frac{s}{2\sigma\sqrt{3}}, \quad \text{for}  \quad 0< s\leq \frac{4\sigma}{\sqrt{3}}.
\end{align}
This inequality demonstrates an inverse relationship between  $\sigma$ and the probability associated with the interval $\mu \pm s/2$, where $s$ is the interval length. This probability is referred to as the  \textit{Dispersion Probability} (DP), has a bound that depends on $\sigma$ and can be used to quantify the uncertainty of $X$. For example, if $X\sim\mathcal{N}(\mu,\sigma^2)$, then: 
\begin{align}
    P\left(|X-\mu|<\frac{s}{2}\right)&=\Phi\left(\frac{s}{2\sigma}\right)-\Phi\left(-\frac{s}{2\sigma}\right)=2\left(\Phi\left(\frac{s}{2\sigma}\right)-\frac{1}{2} \right)\\
    &=2\Phi\left(\frac{s}{2\sigma}\right)-1, \nonumber
\end{align}
where $\Phi$ is the \textit{Cumulative Density Function} (CDF) of the standard Gaussian variable.
In this identity, the monotonically decreasing relationship between probability and uncertainty (represented by $\sigma$) is once again observed. 

\begin{figure}[tb] 
    \centering
    \includegraphics[width=0.3\textwidth]{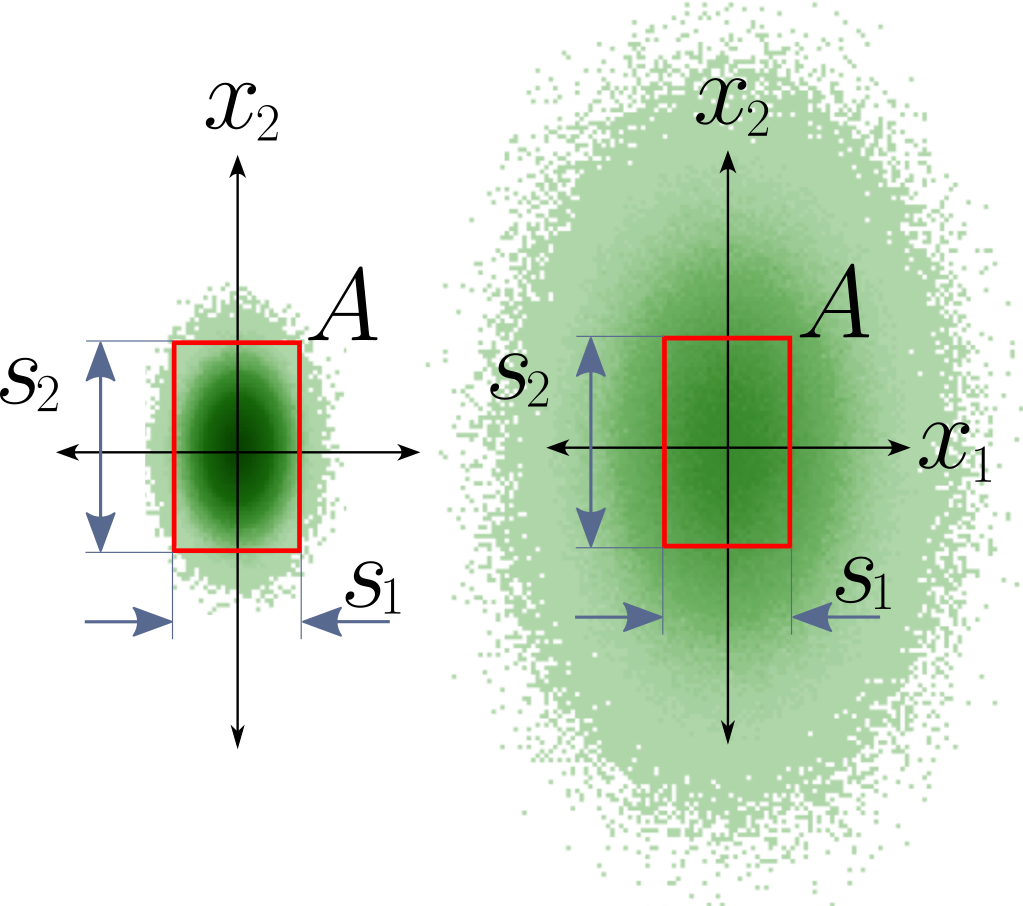}
    \caption{Two histograms of different 2D Gaussian variables are shown. The variance of the variable on the left is lower than that of the variable on the right. Above these histograms, the same integration region $A$ is displayed. }
    \label{fig:hyperrectangle}
\end{figure}

In the multivariate case, with $\mathbf{X}\in\mathbb{R}^N$, $\mathbf{X}\sim\mathcal{N}(\mathbf{\mu},\mathbf{\Sigma})$, and A is the hyperrrectangle with sides $s_i$ given by: 
\begin{equation}
A=\{(x_1,x_2,\ldots,x_N)\in\mathbb{R}^N:| x_i-\mu_i |<s_i/2,\ \forall i=1,\ldots,N\},
\label{def_A}
\end{equation}
the DP of $\mathbf{X}$ is:
\begin{align}
    \label{eq:exploration_map}
    P(\mathbf{X}\in A)=\int_A\mathcal{N}(\mathbf{\mu},\mathbf{\Sigma})(w)dw.  
\end{align}
A graphical representation of this concept is shown in Fig.~\ref{fig:hyperrectangle}. Histograms of the distribution of the Gaussian random variable $(X_1,X_2)\sim\mathcal{N}(\mathbf{\mu},\mathbf{\Sigma})$ are shown on  both the left and right sides of the figure. For given $(s_1,s_2)$ lengths, the interval $A$ is defined for both histograms. On the left side, the dispersion of the variables is lower than on the right side, making it more likely for $(X_1,X_2)$ to fall into $A$ compared to the histogram on the right side. This higher probability is due to the inverse relationship between the dispersion of the variables and the probability of event $A$ occurring. The parameters $(s_1,s_2)$ act as normalization factors that enable the comparison of dispersion across different magnitudes.
For practical implementations, choosing  \(s_i = \sigma_{i,\text{min}}\) ensures that the condition specified in (\ref{eq:gauss_desig}) is satisfied for all \(\sigma_i \geq \sigma_{i,\text{min}}\). 
The proposed methodology is based on continuously updating a grid map of DPs as the agent explores the environment and generates an uncertainty map from these probabilities (as described in Sec.~\ref{sec:uncertainty_frontiers}). 

Based on the sensor model and the agent's location, each measurement impacts a specific cell on the DP grid.  The dispersion probability $p_k$ of the $k$-th cell is updated when $\boldsymbol{\mu} \in R_k$, where \(R_k\) represents the region covered by the cell. 

There is a relationship between $\mathbf{\Sigma_k}$ and $p_k$ of a $k$-th cell for a $N$ dimensional estimation vector,  given by:
\begin{align}\label{eq:relationship_p_sigma}
    |\mathbf{\Sigma}_k|^{\frac{1}{2N}}=\tilde{\sigma}_k \geq \frac{a}{p_k^{1/N}},
\end{align}   
where \(a\) is a constant depending on \(N\) and all \(s_i\) parameters, and \(\tilde{\sigma}_k\) is the geometric mean of the standard deviations of the \(k\)-th cell. These results are demonstrated in \ref{ap:uncertainty_mapping_demo}. This relationship establishes a connection between the PD and the D-opt criterion, a standard measure of uncertainty in ASLAM~\cite{carrillo2012comparison}. However, each DP is bounded between 0 and 1 with a clear probabilistic interpretation. In contrast, the D-opt criterion is unbounded and can only be interpreted when the estimation variables represent lengths, such as the volume of the hyperellipsoid defined by \(\mathbf{\Sigma}_k\). Additionally, the DP grid facilitates the identification of poses that significantly influence map coverage and uncertainty (see Section \ref{sec:relative_entropy} for more details). 

\begin{figure}[t] 
    \centering
    \includegraphics[width=0.45\textwidth]{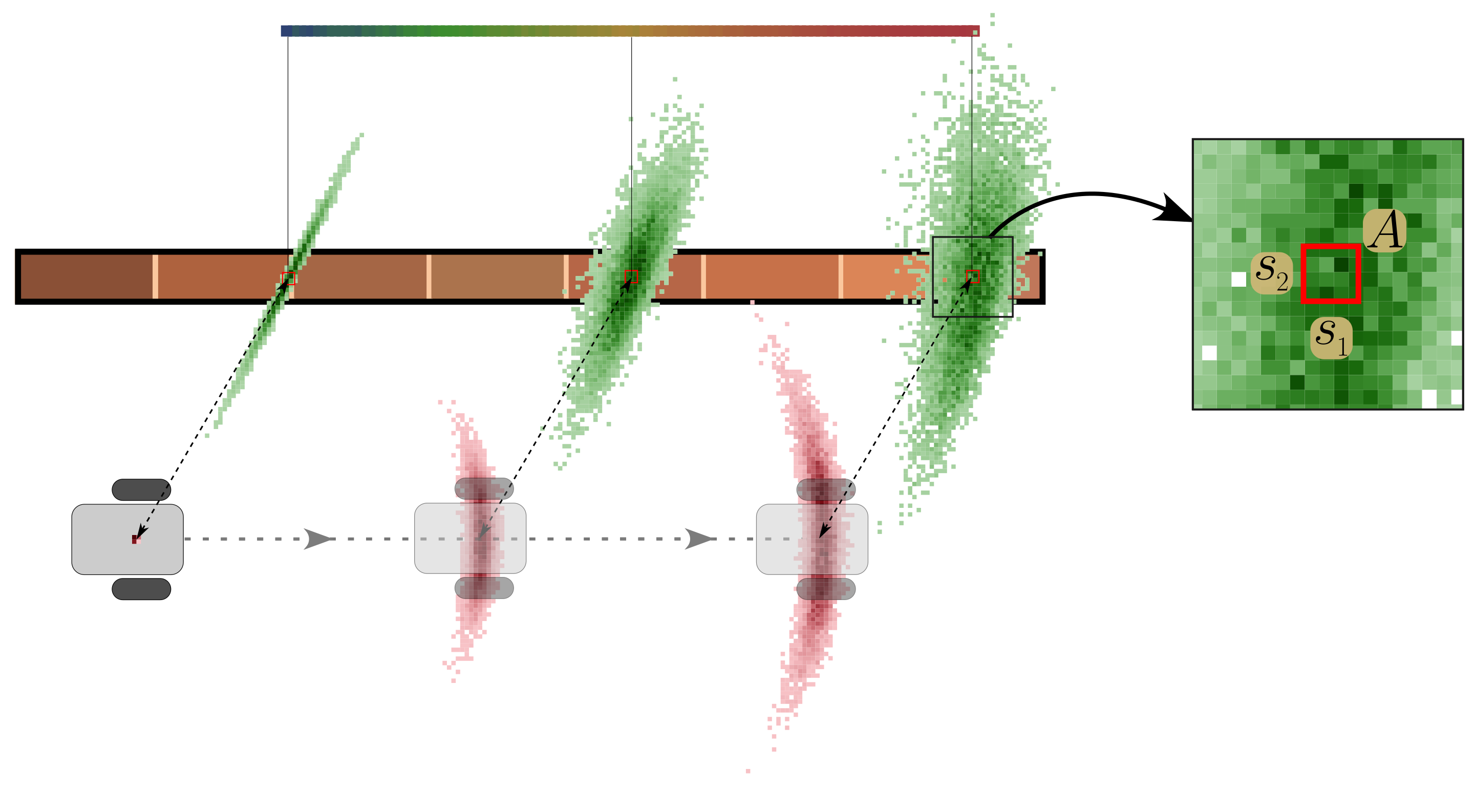}
    \caption{The evolution of the agent's pose uncertainty (represented by red squares) and the measurement from a range sensor, propagated through the agent's uncertainty (represented by green squares), are shown. The intensity of the color indicates the likelihood of the measurement in that region. The color bar at the top shows the $p_k$ values of the cells, with higher $p_k$ values corresponding to bluer colors. }
    \label{fig:curvas_incertidumbre}
\end{figure}

Fig.~\ref{fig:curvas_incertidumbre} illustrates a simple example that identifies patterns of uncertainty evolution. Consider an agent located at a specific position with a degree of uncertainty, equipped with a range sensor to measure distances to objects, such as the distance to a brick wall. 
The agent’s positions, obtained through a Monte Carlo simulation of its odometry accounting for errors, are depicted in varying shades of red. Darker shades of red indicate a higher likelihood of the agent's presence in that region. In the first step of this simulation, the uncertainty of the agent's position is confined to a small area. From this region, an arrow emerges, indicating a measurement made by the sensor, with its dispersion depicted in varying shades of green. Note that the increasing uncertainty in the agent's position propagates to the sensor's measurements in the global frame. 

In this initial step, since the agent's uncertainty is minimal, the measurement uncertainty with respect to the agent is similar to the measurement uncertainty in the global frame.
As the simulation progresses, the increasing position uncertainty of the agent leads to growing measurement uncertainty in the global frame.
Specifically, if measurements are given in polar coordinates $[\rho,\theta]$ relative to the agent's position, itself modeled by the random vector $[x_a, y_a, \phi_a]$ in the global frame, then these measurements are also modeled as a random vector:   
\begin{align}\label{eq:measuring_system}
    \textbf{X}=\begin{bmatrix}x_1\\x_2 \end{bmatrix}=\begin{bmatrix}
        x_a+\rho\cos(\theta+\phi_a)\\
        y_a+\rho\sin(\theta+\phi_a)
    \end{bmatrix}.
\end{align}

Then, given a grid map with cell side \(d\) and:
\[ A = \{(x_1,x_2) \in \mathbb{R}^2 : |x_i - \mu_i| < s_i/2, \ i=1,2 \} \]
for given \(s_1 = s_2 = \sigma_{\text{min}} = 0.1\,m\),
\begin{align}
    P(\mathbf{X} \in A) = P[&\mu_1 - s_1/2 < x_1 < \mu_1 + s_1/2, \nonumber \\
    &\mu_2 - s_2/2 < x_2 < \mu_2 + s_2/2].
\end{align}
In particular, if \(\mathbf{X} \sim \mathcal{N}(\mathbf{\mu}, \mathbf{\Sigma})\), then: 
\begin{align}\label{eq:Pa_approx}
    P(\mathbf{X} \in A) = & F_\mathbf{X}(\mu_1 + s_1/2, \mu_2 + s_2/2) \\
    - & F_\mathbf{X}(\mu_1 - s_1/2, \mu_2 + s_2/2) \nonumber \\
    - & F_\mathbf{X}(\mu_1 + s_1/2, \mu_2 - s_2/2) \nonumber \\
    + & F_\mathbf{X}(\mu_1 - s_1/2, \mu_2 - s_2/2), \nonumber    
\end{align}
where $F_\mathbf{X}$ is the CDF of the random vector $\mathbf{X}$. 
Finally, the probability $p_k$ obtained using (\ref{eq:Pa_approx}) is assigned to the $k$-th cell if $(\mu_1,\mu_2) \in R_k$. Note that to compute \(p_k\), an estimation of \(\mathbf{\mu}\) and \(\mathbf{\Sigma}\) for each measurement is required. The previous example (Fig.~\ref{fig:curvas_incertidumbre}) calculates this using the variance-covariance matrix and the vector of sample means at each simulation step. In practical applications, integrating a SLAM subsystem is essential to provide accurate localization and associated uncertainty for the agent. The color bar at the top of Fig.~\ref{fig:curvas_incertidumbre} visually indicates the levels of measurement uncertainty in the global frame, where blue signifies lower uncertainty and red denotes higher uncertainty. 

\begin{figure}[t] 
    \centering
    \includegraphics[width=0.45\textwidth]{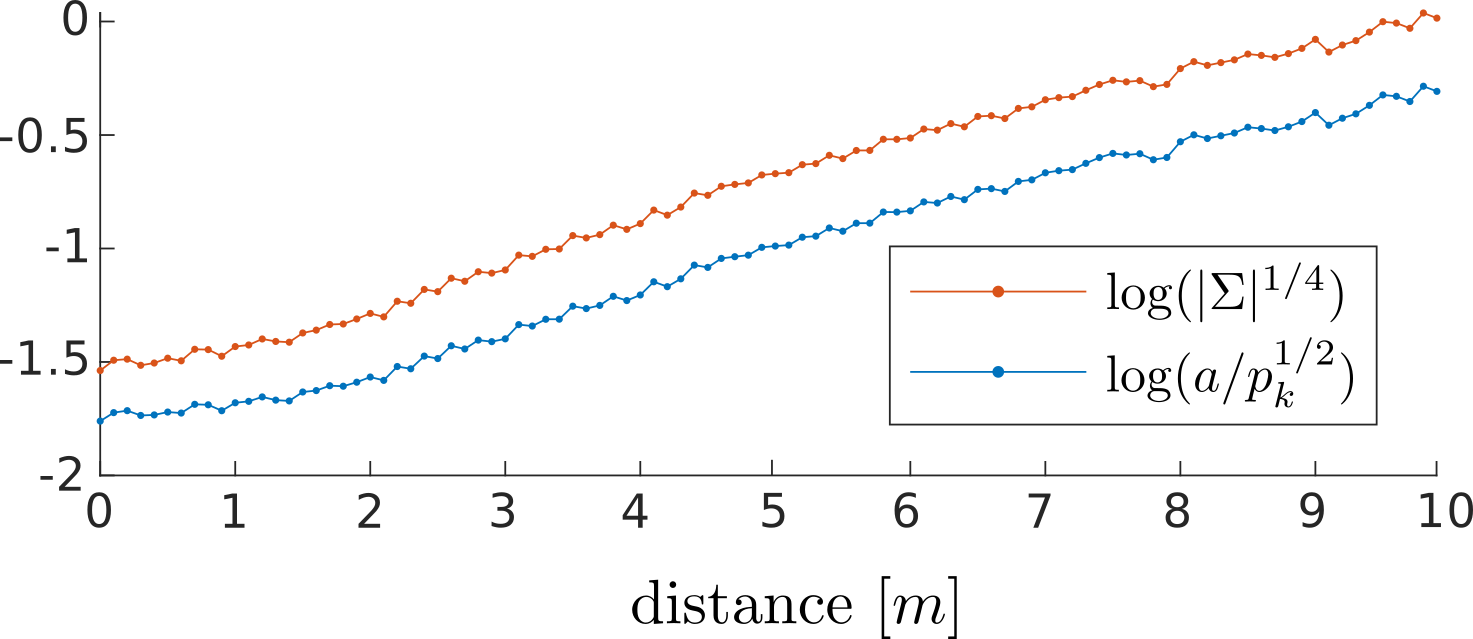}
    \caption{The evolution of the uncertainty measure is shown. The logarithm of the terms from (\ref{eq:relationship_p_sigma}) is plotted for visualization purposes in the 2D case. }
    \label{fig:curvas_p_sigma}
\end{figure}

The relationship described by  (\ref{eq:relationship_p_sigma}) is illustrated in Fig.~\ref{fig:curvas_p_sigma}, along with the DP calculated according to  (\ref{eq:Pa_approx}). To enhance visualization, the logarithms of these variables are plotted.  In this example, both the determinant-based measure and the proposed measure exhibit similar behaviors, aligning with the broader principle that a DP grid is suitable for representing measurement uncertainties.

In practice, many probabilities are assigned to the same cell, making iterative fusion necessary for this information as data is acquired. The most commonly employed method for this purpose is the Bayesian updating algorithm known as log-odds. Modifications and implementation details of this method are discussed in  Sec.~\ref{sec:log-odds}. 

\subsection{Cells update using log-odds}
\label{sec:log-odds}
Each measurement involves a Bayesian update of the DP of the corresponding cell.
As these probabilities represent the probability of the binary event $A$ (see(~\ref{def_A})), a log-odds update strategy can be implemented.
In recent years, methods such as those proposed by~\cite{thrun2003learning,thrun2001learning} have enabled efficient updating of grid maps using the log-odds method.
Among all the existing proposals, \textit{Octrees}~\cite{hornung2013octomap} stands out for 3D grid maps. 

The idea behind these methods is to use the log-odds representation of the occupancy probability of each cell given by:
\begin{align}\label{eq:log_odds_old}
    \ell_{t,k}=\ell_{t-1,k}+\ell_{\text{new}},
\end{align}
with
\begin{equation}
    \ell_{t,k}=\ln\frac{p_{t,k}}{1-p_{t,k}}  \  \Leftrightarrow  \    p_{t,k}=1-\frac{1}{1+e^{\ell_{t,k}}},
\label{eq:log_odds_prob_p2l}
\end{equation}
where $p_{t,k}$ is the probability stored in the $k$-th cell at time $t$, and $\ell_{\text{new}}$ represents the log-odds of the new information obtained by the agent about the $k$-th cell.

The main issue from this perspective is the assumption that poses are known, which is not the case in reality. When dealing with unknown poses,  reversing a large and erroneous map state becomes challenging, requiring either a large value of $|\ell_{\text{new}}|$ or multiple consistent observations to correct $|\ell_{t,k}|$. This issue often arises during extended exploratory trajectories when small estimation errors accumulate significantly, increasing localization uncertainty. Returning to previously visited areas enables SLAM systems to perform loop closures and rectify map inconsistencies. Therefore, the update rule should be flexible enough to accommodate sudden changes in the map. The log-odds map representation has been improved in works~\cite{hornung2013octomap, agha2019confidence} by introducing thresholds that bound the log-odds values between $\ell_{\text{max}}=3.5$ and $\ell_{\text{min}}=-2.0$. For computational purposes, only two values are typically used for updating cells: $\ell_{\text{occ}}=0.85$ or $\ell_{\text{free}}=-0.4$. In other words, $\ell_{\text{new}}$ can only be either $\ell_{\text{free}}$ or $\ell_{\text{occ}}$. 

This update rule continues to produce a map with many saturated cells because the recursive application of (\ref{eq:log_odds_old}) depends on the frequency of cell updates rather than the sensor's uncertainty. Therefore, this map representation is not able to accurately represent sensor uncertainty~\citep{stachniss2003exploring}. To address this issue, a new update rule is proposed to prevent saturation and better represent uncertainty. The cell update formula (\ref{eq:log_odds_old}) is modified as follows: 
\begin{align}\label{eq:log_odds}
    \ell_{t,k}=\ell_{t-1,k}+\kappa\left(\ell_{\text{new}}-\ell_{t-1,k}\right).
\end{align}
With this modification $\ell_{t,k}$ converges close to $\ell_{\text{new}}$ and does not reach the saturation limits $\ell_{\text{max}}$ or $\ell_{\text{min}}$. 
The value of $\ell_{\text{new}}$ can be obtained by integrating the \textit{probability density function} (PDF) over the rectangular region $A$, such as  (\ref{eq:exploration_map}) for the Gaussian case, and then using (\ref{eq:log_odds_prob_p2l}). The design parameter $\kappa \in [0,\  1]$ is set to $1/2$, which regulates the convergence rate of  $\ell_{t,k}$ to $\ell_{\text{new}}$. 


In this context, as the agent explores, increased uncertainty can trigger situations where a cell that was considered “explored” becomes reclassified as “unexplored.” In practical terms, this occurs when the cell's log-odds value crosses below a certain threshold, reversing its occupancy status. To address this issue, we introduce the parameter $\beta$, which defines a DP associated with an unknown cell or a previously established prior for the map. For an environment with no prior knowledge, $\beta$ can be set to reflect the maximum tolerable uncertainty, effectively establishing a boundary between acceptable and unacceptable occupancy states. A numerical example illustrating how to obtain this parameter is provided in \ref{ap:beta}.

Once $\beta$ is defined, the corresponding log-odds $\ell_\beta$ marks the transition between “unexplored” and “explored.” Thus, when  $\ell_{k,t}>\ell_\beta$ changes to $\ell_{k,t}<\ell_\beta$, the map would erroneously classify an already explored cell as unexplored. Such transitions are undesirable because they invalidate prior information. To prevent this, the following adjustment to the update rule is proposed: 
\begin{align}\label{eq:log_odds_exploration}
    \ell_{t,k}=
    \begin{cases}
        \ell_{t-1,k}  & \text{if} \ \ell_{t-1,k}>\max\{\ell_{\beta},\ell_{\text{new}}\}    \\
        \ell_{t-1,k}+\kappa(\ell_{\text{new}}-\ell_{t-1,k}) & \text{otherwise,}
    \end{cases},
\end{align}
where $\ell_{\beta}$ is obtained using (\ref{eq:log_odds_prob_p2l}) with $p_{t,k}=\beta$.
By considering these update rules, the DP 
 does not grow over areas already explored with lower uncertainty.
Moreover, navigating with high uncertainty over explored areas has no impact on the DP grid.
Finally, the uncertainty only increases in cells that change from unexplored ($\ell_{t,k}=\ell_\beta$) to explored with high uncertainty ($\ell_{t,k}<\ell_\beta$). 

When an object is sensed, the measurement not only includes the distance to the object but also classifies the space between the object and the sensor as free. 
Considering this, and assuming that the sensor, such as a LiDAR or a camera, provides multiple measurements, the agent can simultaneously sense a region of the map, known as the Field of View (FOV). These considerations allow extrapolating the ideas constructed when analyzing the particular example in Sec.~\ref{sec:exploration_map}. 

\subsection{Uncertainty Map and Uncertainty Frontiers}
\label{sec:uncertainty_frontiers}
Up to this point, DPs and uncertainties have been somewhat treated as synonymous; however, in reality,  there is only a monotonically decreasing relationship between them.
This section will study this transformation in order to work in units of uncertainty. 
The input of the transformation is the DP grid obtained in the previous section.
For simplicity, the time subscript will be omitted ($p_{t,k}\rightarrow p_k$).
The outcome of the transformation is a grid map called \textit{Uncertainty Map} (UM), where the value in the $k$-th cell is denoted with $U_k$.
This map is constructed from the DP grid developed in Sec.~\ref{sec:exploration_map} using the right side of inequality (\ref{eq:relationship_p_sigma}), explicitly:
\begin{align}\label{eq:uncertainty_map}
    U_k=\frac{\left(\prod_i^N s_i\right)^{1/N}/2\sqrt{3}}{p_k^{1/N}}=\frac{a}{p_k^{1/N}}.   
\end{align}

Due to nonlinearities in the measurement system, abrupt discontinuities can appear in the UM, which  can be smoothed out by revisiting the corresponding map regions.
Based on these considerations, it is proposed to name these discontinuities  \textit{Uncertainty Frontiers} (UF) and use them as exploration objectives.
The UFs are calculated by thresholding the spatial gradient of the UM with a parameter $T_h$, where a cell is considered as a UF only if $||\nabla U_k||_{2}>T_h$.
The threshold $T_h$ is a design parameter that determines the maximum uncertainty jump expected for a given objective.
The smaller this value is, the more sensitive the system becomes, leading to the generation of multiple unnecessary exploration objectives.
The value of $T_h$ depends on the measurement system and the chosen $\sigma_\text{max}$ (and therefore on the $\beta$ obtained, producing an associated $U_\beta$).
Note that if $T_h>U_\beta$, no UF will be generated. 

UFs close to an obstacle are discarded, as they are attributed to an occlusion phenomenon.
Furthermore, any frontiers generated by cells where  $U_k\geq U_\beta$  are also discarded. This consideration keeps bounded the overall measurement uncertainty. 

\subsection{Information measure: Relative entropy}\label{sec:relative_entropy}

Shannon entropy is commonly used to quantify the total information content of a map, as suggested in~\cite{stachniss2005information}. However, it presents certain limitations: unexplored areas modify entropy, and the map's resolution affects the resulting values~\cite{blanco2008novel}. Moreover, Shannon entropy is not well-defined for continuous distributions. 

A similar method for comparing two maps is the~\textit{Kullback-Leibler} divergence.
For the case where measurements in the cells are continuous, this measure is defined as the sum of the relative entropies of each cell in the map, ie.:
\begin{align}\label{eq:information_gain}
D(\mathbf{m}||\mathbf{m'})&=\sum_{k} D_{KL}(P_k ||Q_k) \\
&=\sum_{k}^K  \int_{-\infty}^\infty P_k(w)\ \ln\left(\frac{P_k(w)}{Q_k(w)}\right)\ dw, \nonumber
\end{align}
where $P_k$ is the \textit{probability density function} (pdf) of $k$-th cell of map $\mathbf{m}$ and $Q_k$ is the pdf of the map $\mathbf{m'}$.
It is necessary to know or estimate the \textit{pdfs} involved in (\ref{eq:information_gain}).

Inspired by $D_{KL}$, the \textit{Signed Relative Entropy} (SiREn) is proposed as a measure of the information content of a map, and it is defined as:
\begin{align} \label{eq:siren}
    D_s(\mathbf{m})=\sum_{k} C_k D_{KL}(P_k||Q_k)\text{sgn}\left(|\mathbf{\Sigma}_{Q_k}|-|\mathbf{\Sigma}_{P_k}|\right),
\end{align}
where $C_k$ is the space occupied by the $k$-th cell, $\mathbf{\Sigma}_{P_k}$ and $\mathbf{\Sigma}_{Q_k}$ are the covariance matrices of $P_k$ and $Q_k$ respectively, and $\text{sgn}(x)$ is the sign function. 

In this scenario, the measurement uncertainty is assumed to arise from Gaussian error propagation. Consequently, each \textit{pdf} $P_k$ and $Q_k$ is modeled as a Gaussian distribution with the same mean but different covariance matrices. As a result, $Q_k \sim \mathcal{N}(0,\mathbf{\Sigma}_{Q_k})$ is taken to represent the maximum allowable uncertainty, whereas $P_k \sim \mathcal{N}(0,\mathbf{\Sigma}_{P_k})$ describes the current uncertainty. Although $Q_k$ and $P_k$ are selected as Gaussians, the method can be generalized to any distribution. A crucial requirement is to define a criterion that distinguishes whether the current measurement uncertainty exceeds the maximum allowable threshold. This distinction is essential, as the sign function is employed to penalize instances where the uncertainty surpasses the defined limit, thereby discouraging measurements that fall outside acceptable bounds.

The SiREn has the following properties:
\begin{itemize}
    \item If the entire map is unexplored, then $P_k=Q_k$ and  \mbox{$D_s(\mathbf{m})=0$}. In this case, the unexplored cells do not contribute to the overall relative entropy of the map.
    \item The contribution of each cell to the relative entropy is proportional to the space it occupies, allowing its application to methods based on variable-resolution maps, such as OctoMap \cite{hornung2013octomap}.    
    \item If the measurement uncertainty exceeds the permissible limit, the relative entropy of the map decreases.  This property enables the penalization of trajectories that produce measurements exceeding the acceptable threshold.    
\end{itemize} 

\begin{figure}[t] 
    \centering
    \includegraphics[width=0.4\textwidth]{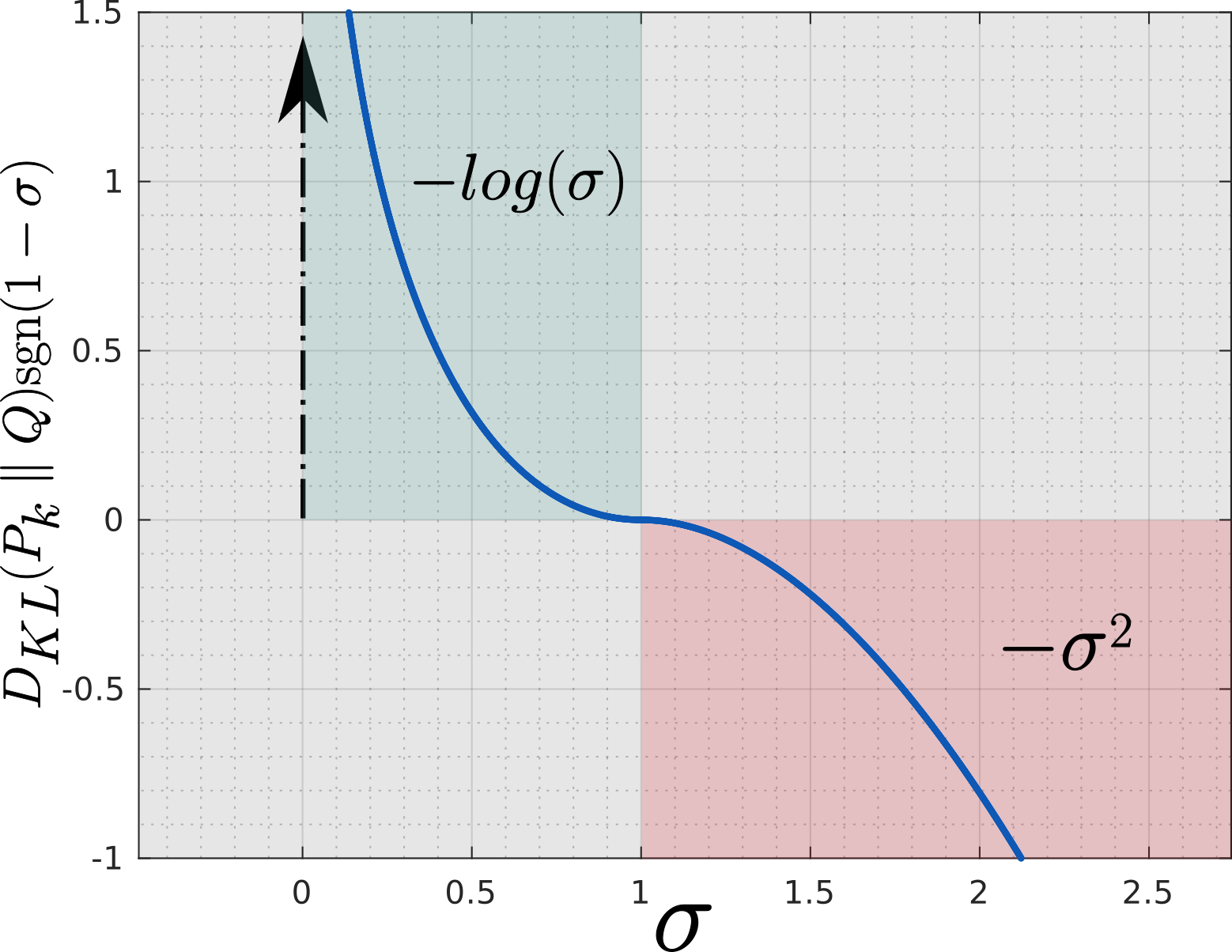}
    \caption{ Behavior of $D_{KL}(P_k||Q_k)$ as $\sigma$ changes is shown. }
    \label{fig:SiREn_vs_sigma}
\end{figure}

For illustrative purposes, one term of the sum (\ref{eq:siren}) is calculated. If $Q_k=Q\sim\mathcal{N}(0,\sigma_\text{max}^2)$ with $\sigma_\text{max}^2=1$ and $P_k\sim\mathcal{N}(0,\sigma^2)$, where $\sigma$ is treated as a variable, then the $k$-th term becomes:
\begin{align}
    D_{KL}(P_k||Q)\text{sgn}(1-\sigma) = 
     \begin{cases}
        -\ln\sigma +\frac{1}{2}\left( \sigma^2-1\right) & \text{if} \ \sigma<1 \\
        \ln\sigma -\frac{1}{2}\left( \sigma^2-1\right) & \text{if} \ \sigma\geq 1.
    \end{cases}  
\end{align}
This function is plotted in Fig. \ref{fig:SiREn_vs_sigma}. 
There is a nonlinear, monotonically decreasing relationship between $D_{KL}(P_k \parallel Q)$ and $\sigma$. Additionally, a vertical asymptote exists at $\sigma = 0$. Moreover, when $\sigma_\text{max} < \sigma$, the resulting value is negative, and for the same distance from the maximum standard deviation, the relative entropy is higher when $\sigma_\text{max} > \sigma$ than when $\sigma_\text{max} < \sigma$. This indicates that SiREn is more sensitive to variations in $\sigma$ before reaching the maximum uncertainty. 

The sensitivity behavior of SiREn demonstrates highly non-linear characteristics, significantly rewarding exploration under low uncertainty compared to the smaller penalties it imposes for high uncertainty exploration. This sensitivity to low uncertainty exploration enables the detection of minimal changes in the map's final uncertainty. In general, the higher the $\sigma_\text{max}$ selected, the less sensitive SiREn becomes to measurement uncertainty. 

Regarding the maximum distance covered in a dead reckoning scenario before SiREn declines, it depends on the characteristics of the uncertainty estimation system. Moreover, with the same system, when various $\sigma_{\max}$ values are employed, the maximum distance before SiREn declines is extended as $\sigma_{\max}$ increases. This feature is valuable for planning tasks, as $\sigma_{\max}$ serves to control the exploration-exploitation trade-off.
Despite this, $\sigma_{\max}$ is not a hard limit for the planning system, because SiREn allows exploration to continue when $\sigma>\sigma_{\max}$ with a penalty that depends on the difference between $\sigma$ and $\sigma_{\max}$. 

As the distributions are assumed to be Gaussian, there exists a closed-form expression to calculate the terms of (\ref{eq:siren}). From~\cite{hershey2007approximating}, and as discussed in \ref{ap:diver}, it follows:
\begin{align}
    D_{KL}(P_k \parallel Q) = -N \ln \frac{\tilde{\sigma}_k}{\sigma_{\text{max}}} - \frac{N}{2} + \frac{N}{2}\left(\frac{\tilde{\sigma}_k}{\sigma_{\text{max}}}\right)^2.
\end{align}
Using the PD approach, the following approximation is obtained:
\begin{align}\label{eq:approx_siren}
    D_{KL}(P_k \parallel Q) \geq \ln \frac{p_k}{\beta} - \frac{N}{2} + \frac{N}{2}\left(\frac{\beta}{p_k}\right)^{\frac{2}{N}}.
\end{align}
From this point forward, \textit{SiREn} and \textit{relative entropy} will be used interchangeably to enhance clarity in the text. 

\section{Computational Cost Analysis} \label{sec:computational_cost}
In this section, the computational cost of the proposed methods is analyzed. To compute the DP measure in (\ref{eq:exploration_map}) or for any other distribution, one can either use a precomputed table for known distributions or perform a \textit{Monte Carlo} (MC) simulation. In the worst-case scenario, computing a single measure requires \(\mathcal{O}(k \, d^3)\)  operations for MC simulation, where \(k\) is the number of samples and \(d\) is the dimension of the covariance matrix. Then, for each cell  affected by this measure, the log-odds update procedures from (\ref{eq:log_odds_prob_p2l}) and (\ref{eq:log_odds_exploration}) must be executed, which incurs a cost of \(\mathcal{O}(N)\). Here, \(N\) is the number of cells modified by the sensor measurement, a value that depends on the sensor's FoV. If a distinct uncertainty measure is computed for each cell (as in the case of cameras), the overall computational cost becomes  \(\mathcal{O}(N \, k \, d^3)\).

On the other hand, the computational cost of the ray-casting method depends on its specific implementation. In a graph-based structure, it can be as low as \(\mathcal{O}(\log N)\), but this will vary depending on how the environment is discretized and queried.

To compute the SiREn measure, (\ref{eq:log_odds_prob_p2l}) and (\ref{eq:approx_siren}) are applied to all cells in the map, denoted by \(\mathbb{N}\). For the UF measure on a DP grid of \(\mathbb{N}\) cells, it is first necessary to compute \(U_k\) using (\ref{eq:log_odds_prob_p2l}) and then (\ref{eq:uncertainty_map}) for each cell. Finally, obtaining UF requires calculating the gradient of \(U_k\). All of these steps have a cost of \(\mathcal{O}(\mathbb{N})\).

The computational cost can also be reduced by storing the map in a hierarchical data structure such as an octree, which would limit the number of required operations to those cells occupying the updated regions.

In conclusion, the worst-case computational cost of the proposed methods is \(\mathcal{O}(N)\), with \(N\) depending on the field of view of the sensor. Computing SiREn and UF over the entire map has a complexity of \(\mathcal{O}(\mathbb{N})\). These costs are comparable to classical methods used in active SLAM systems.

\section{Evaluation}\label{sec:experiments} 

This section aims to demonstrate the behavior of the proposed map model, relative entropy, and uncertainty frontiers through two exemplary experiments, each with its specific objectives. Additionally, it seeks to showcase the proposed set of tools and their performance in various scenarios by implementing a complete ASLAM system, which includes a SLAM system, a path planner, a low-level robot controller, and a high-level trajectory tracking controller, among other subsystems. However, since the primary focus is to assess the behavior of a specific part of the ASLAM system, the rest of the subsystems are kept as simple as possible to isolate the cause-effect relationship of the proposed modifications. 

It is important to clarify that the primary objective of this work is to develop and present methodologies aimed at enhancing specific aspects of Active-SLAM systems, rather than delivering a full-scale Active-SLAM implementation or optimizing trajectory planning in complex environments. Consequently, the use of simplified experimental environments is a deliberate choice to isolate and analyze the cause-effect relationships between key variables without the interference of additional complexities associated with geometric intricacy and real-world navigation constraints.

The first experiment illustrates the behavior of UF, detailing its generation process and its dependence on the parameter $T_h$. 
The second experiment explores the relationship between relative entropy, the uncertainty of the map, and its coverage. For this purpose, four path planners will be compared through simulation: two utilize the Yamauchi or \textit{Classical Frontiers} (CF) \cite{yamauchi1997frontier} , and the other two employ the UF. 

All experiments are performed with a differential robot simulated in Gazebo. 
Besides, a 2D LiDAR mounted on the robot measures distances up to $5m$ with $360^o$ coverage.
The configuration of the uncertainty-aware planning system is described in \ref{sec:planning}.
The KF-SLAM system, used to obtain estimates of each pose with its corresponding variance-covariance matrix, is detailed in  \ref{sec:kf_slam}. 

\subsection{Frontiers test}
\begin{figure}[t] 
    \centering
    \includegraphics[width=0.42\textwidth]{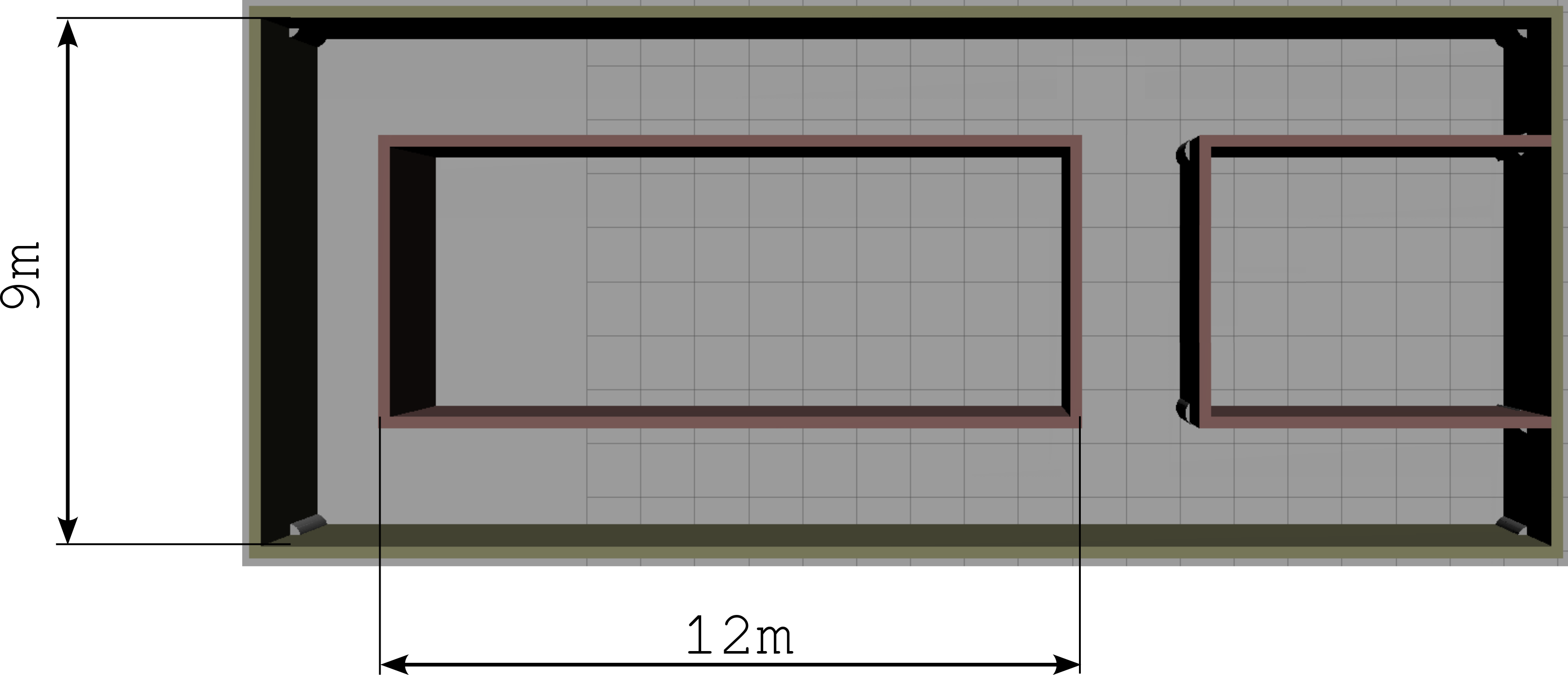}
    \caption{ Map proposed to test UF. Corridors have a width of $2m$ and a length up to $12m$. }
    \label{fig:corridor1}
\end{figure}

\begin{figure}[t] 
    \centering
    \includegraphics[width=0.45\textwidth]{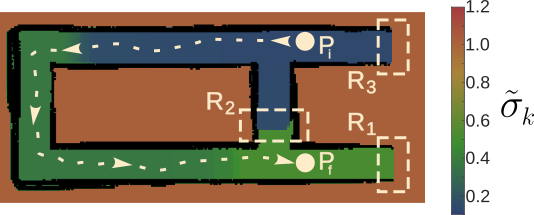}
    \caption{Map proposed to show the UF. The blue zone has less uncertainty than the green zone. The dotted line show the path followed by the robot. }
    \label{fig:corridor2}
\end{figure}

The first simulation aims to demonstrate the benefits of using UF over the CF approach.
The robot is positioned in a building as shown Fig.~\ref{fig:corridor1}.
The robot moves from $P_i$ to $P_f$ by teleoperation, following the path indicated with the dotted line in Fig.~\ref{fig:corridor2}. 
When the robot starts moving, and when it reaches $P_f$, the FOV of LiDAR does not detect the walls of the ending corridor on the right. 
 The colors in the figure represent uncertainty levels stored in the UM, where blue zones indicate less uncertainty than green zones.
Both CF  and UF  that are less than $0.5$m from an obstacle are discarded.
The parameters $\sigma_{\max}=1.0$m, $s_1=s_2=0.1$m and $T_h=0.2$m are set. 

The highlighted three regions, $R_1$, $R_2$, and $R_3$, are recognized as UF following the procedure described in Sec.~\ref{sec:uncertainty_frontiers}.
The average gradients of these regions are $0.28$m, $0.36$m, and $0.45$m, respectively. If the threshold selected is greater than $0.28m$, then $R_1$ will not be considered as a UF. According to the procedure proposed in \cite{yamauchi1997frontier}, only $R_1$ and $R_3$ are classified as CF. 

Depending on the detected objectives, the planner selects a trajectory. When using the CF method, $R_1$ and $R_3$ are designated exploration objectives. However, when using UF, $R_2$ also becomes an exploration objective, which could lead to loop closure. The planner must determine the path that maximizes information extraction through either exploration or exploitation. When the FOV of the agent reaches a CF, it is eliminated, and potentially new ones are created if no obstacles are present. The situation differs for UFs. A UF is eliminated when $||\nabla U_k||_{2} < T_h$, which occurs when the maximum uncertainty in the frontier cells is reduced. Therefore, the path planning stage must consider the uncertainty of both the robot and the measurements upon reaching a UF. The  \ref{sec:planning} describes a path planner incorporating a UF elimination mechanism. 

In conclusion, unlike CFs, UFs serve as exploration objectives and aim to reduce map uncertainty by closing loops and re-exploring previously explored areas with lower uncertainty. 

\subsection{SiREn and UF test}
This second experiment highlights the benefits of using relative entropy and its connections to other metrics, such as coverage and uncertainty estimation, in map construction and landmark estimation. Additionally, it provides a comparison of planning systems with and without UFs. 

\begin{figure}[b] 
    \centering
    \includegraphics[width=0.4\textwidth]{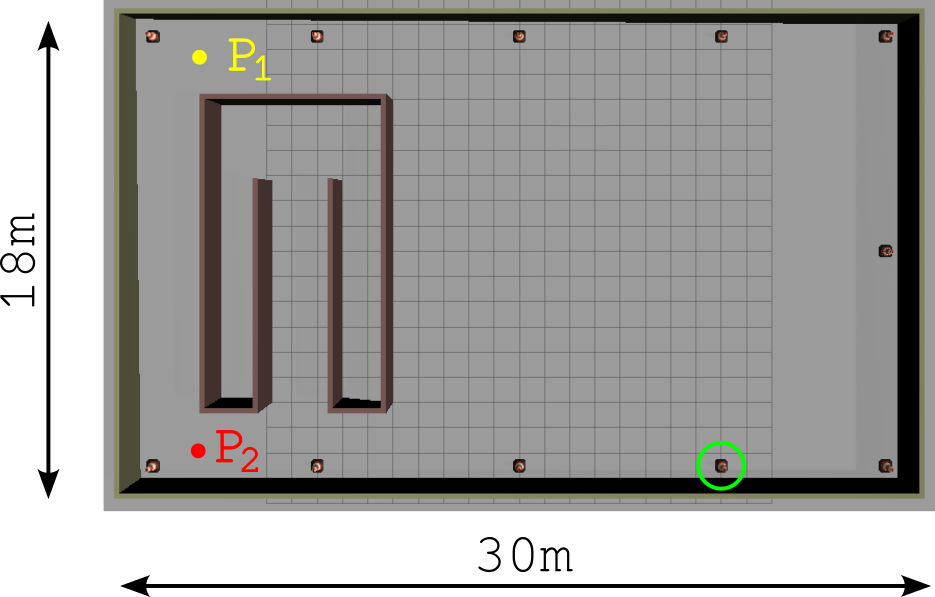}
    \caption{Scenario proposed to test the SiREn measure and UF.  Two possible initial positions of the robot, along with the configuration of the landmarks, are shown. In yellow circle is the initial pose $P_1$ (up), in red  circle the initial pose $P_2$ (down), the red points on the periphery are the landmarks used for \textit{complete} and \textit{incomplete} dispositions. }
    \label{fig:galpon}
\end{figure}

For this, the robot is placed in a rectangular closed space with landmarks arranged on its perimeter, as shown  Fig.~\ref{fig:galpon}. The environment has an open space on the right and corridors on the left. This scenario is designed to evaluate the system's performance in two types of environments, which are simple enough to isolate the cause-effect relationship of the proposed modifications. 
The experimental layouts vary the agent's initial position ($P_1$ and $P_2$ in Fig.~\ref{fig:galpon}) and the number of available landmarks, referenced to as \textit{complete} or \textit{incomplete}. The \textit{complete}  setup includes all perimeter landmarks, while the \textit{incomplete} setup excludes the landmark highlighted in green.

On the other hand, four \textit{path planning systems} (PPS) are proposed for comparison and analysis purposes. The first and simplest system, PPS1, implements CF as exploration objectives and uses an RRT planner to generate the path and execute the shortest trajectory, also known as a greedy planner. The second system, PPS2, is similar to the first but employs a modified RRT*-planner, as described in~\ref{sec:planning}. The third system, PPS3, which is identical to PPS2, uses UFs as exploration objectives with a $\sigma_\text{max}=0.6$m. The fourth and final system, PPS4, is similar to PPS3 but with a $\sigma_\text{max}=1.0$m. 

In all cases, the initial uncertainty of the agent is $\tilde{\sigma}_0=0.1$m and $T_h=0.2$m when UFs are implemented.  As all PPS are based on a random planner, each experiment is repeated at least five times to reduce random effects. As a stopping criterion, all PPS are terminated when no more UFs or CFs are reachable. In total, 16 experiments are performed, each repeated at least five times. The results are presented in the following section. 

\subsubsection{Results and Discussion}\label{sec:results}
\begin{figure}[t] 
    \centering
    \includegraphics[width=0.4\textwidth]{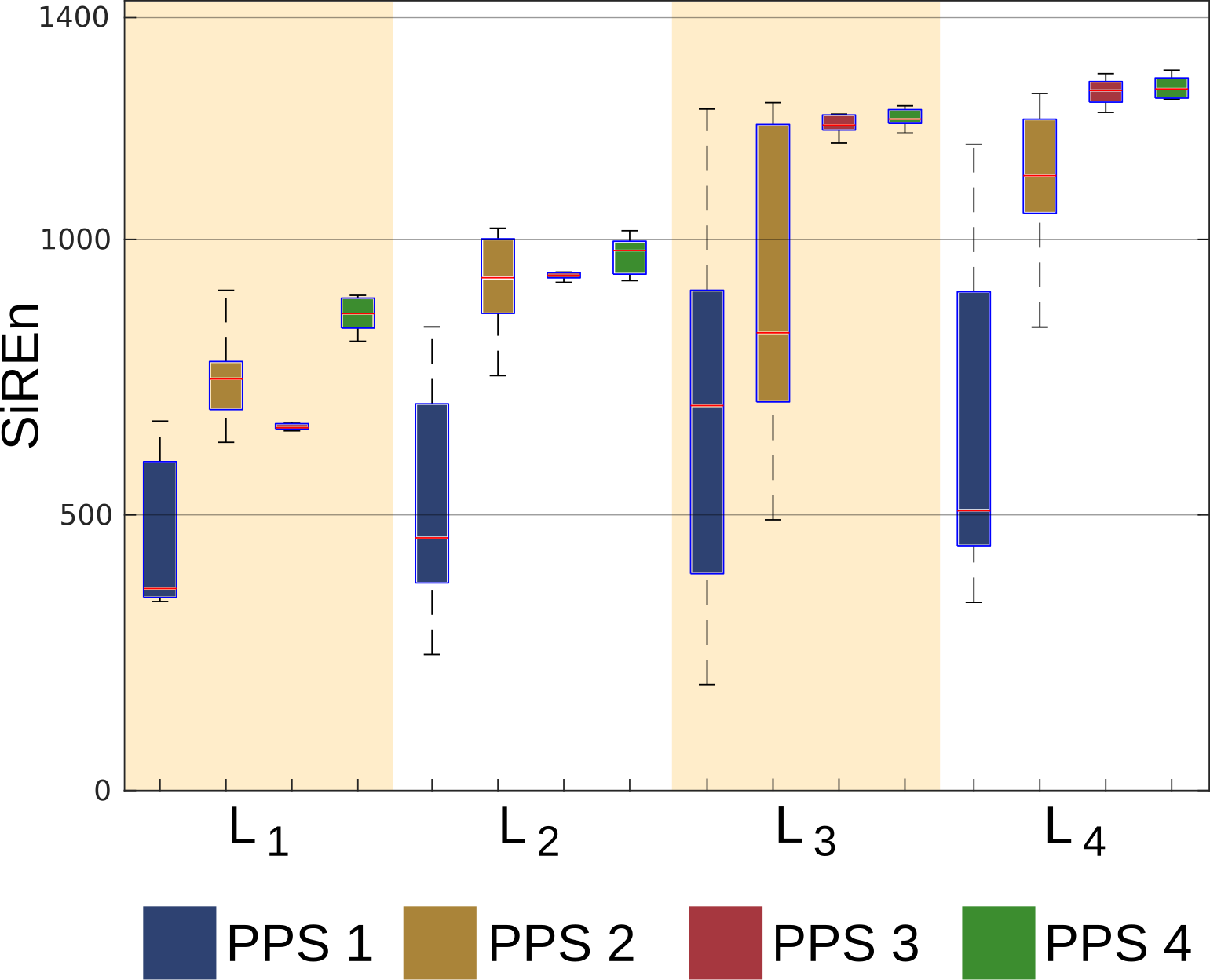}
    \caption{ Boxplots for the SiREn measurements across all experiments. Each layout, denoted by $L_i$, varies in the arrangement of landmarks and initial positions: $L_1$ and $L_2$ feature incomplete layouts with initial positions $P_2$ and $P_1$, respectively. Meanwhile, $L_3$ and $L_4$ feature complete layouts, with $L_3$ starting at $P_1$ and $L_4$ at $P_2$.}        
    \label{fig:all_experiments2}
\end{figure}

In Fig.~ \ref{fig:all_experiments2}, the SiREn values obtained for each experimental setup are displayed in a boxplot. Generally, PPS1 and PPS2 exhibit greater dispersion in the final SiREn values compared to PPS3 and PPS4. Additionally, the median SiREn values for PPS3 and PPS4 exceed those of PPS1 and PPS2, except in the incomplete layout with $\sigma_\text{max}=0.6$m. 

A representative UM of each experiment is presented in Fig.~\ref{fig:all_um} to discuss these results. Starting with PPS1, the entire map is explored in all layouts, but notable variability in measurement uncertainties is observed. This variability stems from PPS1's greedy strategy to reach the CF, which is influenced by the randomness of the RRT. Significant discontinuities in the UM are also observed, as indicated by abrupt color changes. These variations are reflected in the SiREn values obtained, with areas explored under greater uncertainty (near $\sigma_{\max}$) being more extensive than those explored under lower uncertainty (marked in blue). 

Compared to PPS1, PPS2 also covers the entire map but exhibits lower variability in measurement uncertainty. This improvement occurs because PPS2 employs a modified RRT* that accounts for the agent's uncertainty, enabling the planner to choose paths through areas of lower uncertainty that include well-known landmarks. Additionally, while discontinuities are still present in the UM, they are less abrupt and frequent than those observed with PPS1. This behavior is reflected in the SiREn values obtained, as the areas explored with lower uncertainty are more extensive, resulting in higher SiREn values. 

For PPS3, the map is not fully explored when using an incomplete landmark disposition.
 This occurs because PPS3 employs UF as exploration objectives with a~$\sigma_{\max}=0.6$m, leading the path planner to exhaust exploration objectives when the measurement uncertainty approaches this maximum threshold.  Additionally, measurement uncertainty transitions smoothly and continuously, except when obstacles are present, resulting in gradual color changes on the map. PPS3 uses the same path planner as PPS2 but differs in that it utilizes UF for exploration objectives, which explains the observed results. Generally, PPS3 explores areas with lower uncertainty than PPS2, achieving higher SiREn values. However, when operating with an incomplete landmark arrangement, PPS3 explores only part of the map, resulting in fewer cells contributing to the SiREn calculation compared to other scenarios.

Lastly, for PPS4, the map is extensively explored in all layouts, with measurement uncertainty varying homogeneously. Although the path planner also exhausts exploration objectives when uncertainty approaches $\sigma_{\max}=1.0m$, PPS4 covers greater distances due to its higher measurement uncertainty $\sigma_{\max}$. This PPS explores the most extensive area with the lowest uncertainty, resulting in higher SiREn values. Moreover, although it uses the same path planner as PPS2, the critical difference lies in the fact that PPS4 can correct sudden changes in uncertainty in the UM by using UFs as exploration objectives. This adjustment results in higher SiREn values and lower variability, even when employing a path planner with random components. 

Notably, the layout and PPS that achieved the highest SiREn values used a complete landmark disposition, starting from the lower position with PPS4. This is because SiREn, as a non-linear relative entropy measure, rewards exploration with low uncertainty more significantly than it penalizes exploration with high uncertainty. This sensitivity to low uncertainty enables differentiation between starting positions. When starting from the lower position, the uncertainty during exploration in the corridors is lower compared to starting from the upper position, resulting in a higher SiREn. 

For those PPS that utilize UF, the path planner explores the environment in a limited manner. This approach is particularly advantageous for large maps that cannot be fully explored in a single session, especially under challenging conditions such as poor landmark distribution or extensive open spaces.  

On the other hand, to relate the UM to landmark uncertainty, an additional analysis is performed on the maps produced with PPS3 and PPS4. 
The process begins by calculating the median $\tilde{\sigma}_l$ for each landmark across the experiments. This median $\tilde{\sigma}_l$ is derived from the geometric mean of the uncertainty associated with each landmark, calculated as $\tilde{\sigma}_l = |\mathbf{\Sigma}_l|^{1/2N}$, where $\mathbf{\Sigma}_l$ is the variance-covariance matrix for the landmark. After determining the median $\tilde{\sigma}_l$ for each landmark, the next step involves calculating the overall median of these $\tilde{\sigma}_l$ values. This overall median, denoted as $\tilde{\sigma}$, serves as a composite measure of uncertainty for all observed landmarks. It is found that this composite measure correlates strongly, at $97\%$, with the median uncertainty of the entire map, denoted as $\tilde{U}_k$. The relationship between the overall median $\tilde{\sigma}$ of the landmarks and the median uncertainty $\tilde{U}_k$ of the Uncertainty Map is further illustrated through the following fitted curve: 
\begin{figure*}[t] 
    \centering
    \includegraphics[width=0.9\textwidth]{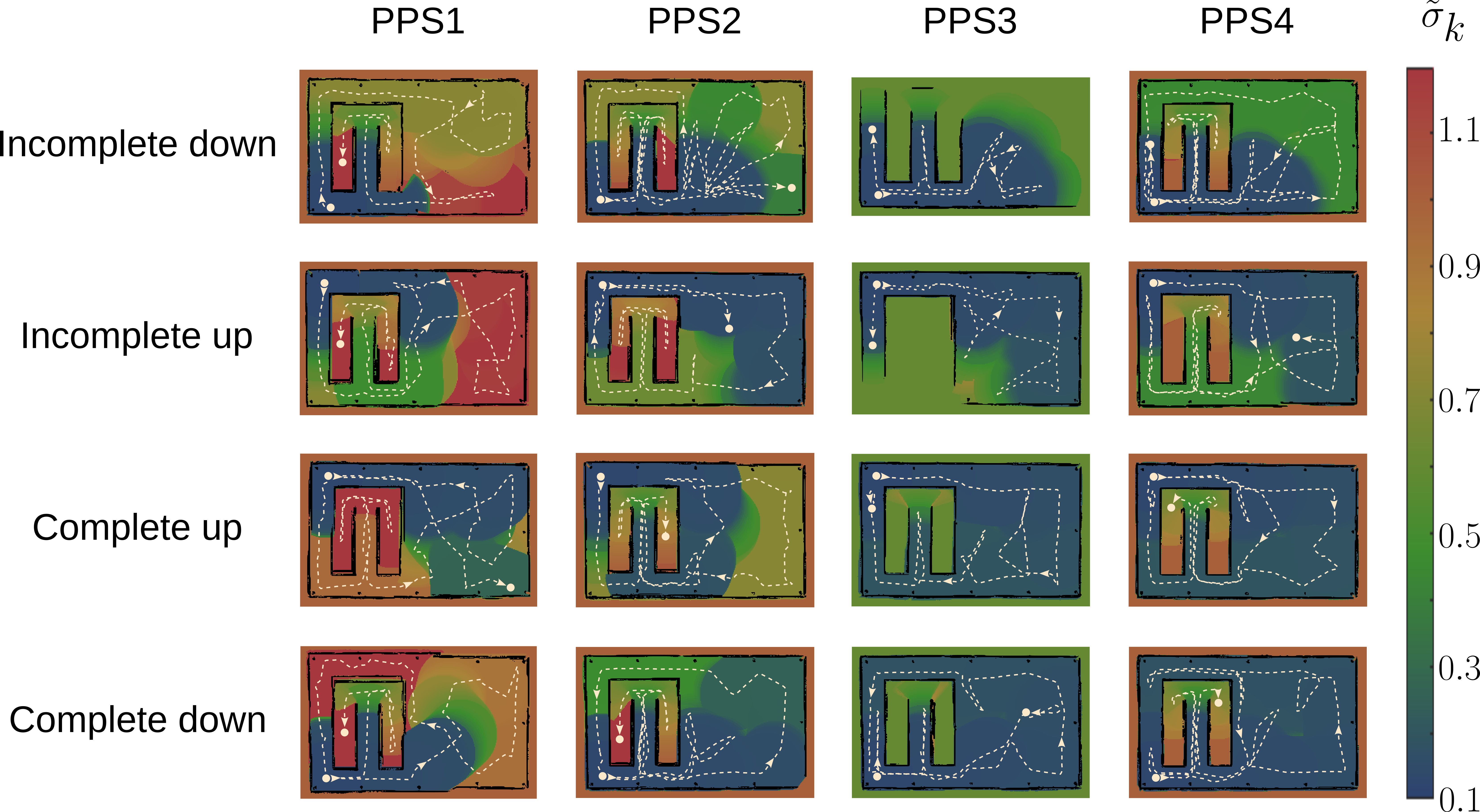}
    \caption{ A sample of each layout and PPS is showed with its trajectory. The color of the cells represents the uncertainty of the measurement accoding to de color bar at the right. The dotted line is the path followed by the robot.}        
    \label{fig:all_um}
\end{figure*}
\begin{align}
    \tilde{U_k}=0.971\tilde{\sigma}+0.00127.
\end{align}
Additionally, the slope is close to 1, and the intercept is near 0, indicating that the map's mean uncertainty closely mirrors the mean uncertainty of the landmarks. In other words, the uncertainty of reconstruction associated with $\tilde{\sigma}$ can be estimated by $\tilde{U}_k$ with negligible error, or vice versa. 

Through the SiREn of each reconstruction, it is possible to establish an order of impact of those parameters that affect the quality of the final reconstruction. Firstly, the arrangement of the landmarks 
has the most significant influence, followed by  the initial position of the agent, then by  the type of PPS used (whether uncertainty-aware or not),  subsequently by the use or not of UF, and finally, by the $\sigma_{\max}$. 
The results are linked to the well-known fact that landmarks' amount and distribution affect the reconstruction process. Additionally, the agent's initial position influences the map's outcome, as a poor initialization can lead to a map with more significant uncertainty. Finally, $\sigma_{\max}$ has the least impact on the signed relative entropy compared to the other parameters. 

Ultimately, while coverage influences relative entropy, higher coverage does not necessarily lead to higher relative entropy since it is highly dependent on measurement uncertainty. 

\section{Conclusion}\label{sec:conclusions}
This work proposes a set of tools designed for ASLAM systems and introduces a new methodology for modeling uncertainty through PDs. Using these, an UM is constructed, from which both the UF and the SiREn index are derived. This approach is independent of the SLAM system used, requiring only uncertainty estimations. Also, any type of sensor, including cameras, LiDARs, or sensor fusion, can be integrated.

Furthermore, integrating this map modeling approach with a UF-based planning system enables the agent to autonomously explore open spaces, a behavior not previously observed in the Active SLAM literature. Additionally, UFs have proven to be an effective tool for determining stopping criterion, addressing an open problem highlighted in the literature. Conversely, in cases where a path planner based on uncertainty is used, the UF can close loops and re-explore areas to improve the map's quality compared to the same system using CF.

It is crucial to note that relative entropy, or SiREn, is sensitive to both coverage and reconstruction uncertainty. This characteristic is essential for comparing different reconstructions and selecting the most appropriate path, thereby effectively achieving Active SLAM's fundamental objectives. This demonstrates that through the SiREn of each reconstruction, it is possible to establish an order of impact of those parameters that affect the quality of the final reconstruction. The order of effects is as follows: the arrangement of the landmarks, followed by the initial position of the agent, then the type of PPS used (whether uncertainty-aware or not),  subsequently by the use or not of UF, and finally, by the $\sigma_{\max}$.

It is worth noting that the aim of this work was to present a new theoretical tool, so experiments were deliberately kept simple in order to test and easily understand the concept.
Future work will focus on developing a planning system that integrates UF and SiREn, making it applicable to both landmark-based and non-landmark-based SLAM systems.

\section{CRediT author statement}
\textbf{Sebastian Sansoni:} Conceptualization, Methodology, Software, Validation, Formal analysis, Investigation, Resources, Data curation, Writing $-$ Original Draft, Writing $-$ Review \& Editing,  Visualization,  Project administration.
\textbf{Javier Gimenez:} Conceptualization, Formal analysis, Writing $-$ Original Draft, Writing $-$ Review \& Editing, Supervision.
\textbf{Gaston Castro:} Conceptualization, Methodology, Software, Validation, Formal analysis, Investigation, Data Curation, Writing $-$ Review \& Editing.
\textbf{Santiago Tosseti:} Conceptualization,  Resources, Writing $-$ Review \& Editing, Supervision, Funding acquisition.
\textbf{Flavio Capraro:}  Writing $-$ Review \& Editing, Supervision, Resources, Funding acquisition.
\section{Acknowledgements}
Sebastian Sansoni and Gastón Castro are funded by a Latin American doctoral fellowship from the Consejo Nacional de Investigaciones Científicas y Técnicas (CONICET) of Argentina.
\section{Declaration of generative AI and AI-assisted technologies in the writing process}
During the preparation of this work the authors used ChatGPT and Github Copilot in order to  enhancing language clarity and readability. After using this tool/service, the authors reviewed and edited the content as needed and takes full responsibility for the content of the publication.
\bibliographystyle{elsarticle-num}
\bibliography{./biblio_2.bib}

\begin{thebibliography}{10}
\expandafter\ifx\csname url\endcsname\relax
  \def\url#1{\texttt{#1}}\fi
\expandafter\ifx\csname urlprefix\endcsname\relax\def\urlprefix{URL }\fi
\expandafter\ifx\csname href\endcsname\relax
  \def\href#1#2{#2} \def\path#1{#1}\fi

\bibitem{khosoussi2014novel}
K.~Khosoussi, S.~Huang, G.~Dissanayake, Novel insights into the impact of graph structure on slam, in: 2014 IEEE/RSJ International Conference on Intelligent Robots and Systems, IEEE, 2014, pp. 2707--2714.

\bibitem{yamauchi1997frontier}
B.~Yamauchi, A frontier-based approach for autonomous exploration, in: Proceedings 1997 IEEE International Symposium on Computational Intelligence in Robotics and Automation CIRA'97.'Towards New Computational Principles for Robotics and Automation', IEEE, 1997, pp. 146--151.
\newblock \href {https://doi.org/10.1109/CIRA.1997.613851} {\path{doi:10.1109/CIRA.1997.613851}}.

\bibitem{stachniss2005information}
C.~Stachniss, G.~Grisetti, W.~Burgard, Information gain-based exploration using rao-blackwellized particle filters., in: Robotics: Science and systems, Vol.~2, 2005, pp. 65--72.
\newblock \href {https://doi.org/10.15607/RSS.2005.I.009} {\path{doi:10.15607/RSS.2005.I.009}}.

\bibitem{leung2006active}
C.~Leung, S.~Huang, G.~Dissanayake, Active slam using model predictive control and attractor based exploration, in: 2006 IEEE/RSJ International Conference on Intelligent Robots and Systems, IEEE, 2006, pp. 5026--5031.

\bibitem{atanasov2015decentralized}
N.~Atanasov, J.~Le~Ny, K.~Daniilidis, G.~J. Pappas, Decentralized active information acquisition: Theory and application to multi-robot slam, in: 2015 IEEE International Conference on Robotics and Automation (ICRA), IEEE, 2015, pp. 4775--4782.
\newblock \href {https://doi.org/10.1109/ICRA.2015.7139863} {\path{doi:10.1109/ICRA.2015.7139863}}.

\bibitem{elfes1989using}
A.~Elfes, Using occupancy grids for mobile robot perception and navigation, Computer 22~(6) (1989) 46--57.
\newblock \href {https://doi.org/10.1109/2.30720} {\path{doi:10.1109/2.30720}}.

\bibitem{thrun2003learning}
S.~Thrun, Learning occupancy grid maps with forward sensor models, Autonomous robots 15~(2) (2003) 111--127.
\newblock \href {https://doi.org/10.1023/A:1025584807625} {\path{doi:10.1023/A:1025584807625}}.

\bibitem{hornung2013octomap}
A.~Hornung, K.~M. Wurm, M.~Bennewitz, C.~Stachniss, W.~Burgard, Octomap: An efficient probabilistic 3d mapping framework based on octrees, Autonomous Robots 34~(3) (2013) 189--206.
\newblock \href {https://doi.org/10.1007/s10514-012-9321-0} {\path{doi:10.1007/s10514-012-9321-0}}.

\bibitem{moorehead2001autonomous}
S.~J. Moorehead, R.~Simmons, W.~L. Whittaker, Autonomous exploration using multiple sources of information, in: Proceedings 2001 ICRA. IEEE International Conference on Robotics and Automation (Cat. No. 01CH37164), Vol.~3, IEEE, 2001, pp. 3098--3103.
\newblock \href {https://doi.org/10.1109/ROBOT.2001.933093} {\path{doi:10.1109/ROBOT.2001.933093}}.

\bibitem{bourgault2002information}
F.~Bourgault, A.~A. Makarenko, S.~B. Williams, B.~Grocholsky, H.~F. Durrant-Whyte, Information based adaptive robotic exploration, in: IEEE/RSJ international conference on intelligent robots and systems, Vol.~1, IEEE, 2002, pp. 540--545.
\newblock \href {https://doi.org/10.1109/IRDS.2002.1041446} {\path{doi:10.1109/IRDS.2002.1041446}}.

\bibitem{stachniss2003exploring}
C.~Stachniss, W.~Burgard, Exploring unknown environments with mobile robots using coverage maps, in: IJCAI, Vol. 2003, 2003, pp. 1127--1134.
\newblock \href {https://doi.org/10.1109/IROS.2003.1250673} {\path{doi:10.1109/IROS.2003.1250673}}.

\bibitem{o2012gaussian}
S.~T. O’Callaghan, F.~T. Ramos, Gaussian process occupancy maps, The International Journal of Robotics Research 31~(1) (2012) 42--62.
\newblock \href {https://doi.org/10.1177/0278364911421039} {\path{doi:10.1177/0278364911421039}}.

\bibitem{placed2022enough}
J.~A. Placed, J.~A. Castellanos, Enough is enough: Towards autonomous uncertainty-driven stopping criteria, IFAC-PapersOnLine 55~(14) (2022) 126--132.
\newblock \href {https://doi.org/10.1016/j.ifacol.2022.07.594} {\path{doi:10.1016/j.ifacol.2022.07.594}}.

\bibitem{placed2023survey}
J.~A. Placed, J.~Strader, H.~Carrillo, N.~Atanasov, V.~Indelman, L.~Carlone, J.~A. Castellanos, A survey on active simultaneous localization and mapping: State of the art and new frontiers, IEEE Transactions on Robotics (2023).
\newblock \href {https://doi.org/10.1109/TRO.2023.3248510} {\path{doi:10.1109/TRO.2023.3248510}}.

\bibitem{aguirre2023}
E.~Aguirre-Zapata, H.~Alvarez, C.~V. Dagatti, F.~Di~Sciascio, A.~N. Amicarelli, Parametric interpretability of growth kinetics equations in a process model for the life cycle of lobesia botrana, Ecological Modelling -~(-) (2023) --.
\newblock \href {https://doi.org/10.1016/j.ecolmodel.2023.110407} {\path{doi:10.1016/j.ecolmodel.2023.110407}}.

\bibitem{nauck2003measuring}
D.~D. Nauck, Measuring interpretability in rule-based classification systems, in: The 12th IEEE International Conference on Fuzzy Systems, 2003. FUZZ'03., Vol.~1, IEEE, 2003, pp. 196--201.
\newblock \href {https://doi.org/10.1109/FUZZ.2003.1209361} {\path{doi:10.1109/FUZZ.2003.1209361}}.

\bibitem{carrillo2018autonomous}
H.~Carrillo, P.~Dames, V.~Kumar, J.~A. Castellanos, Autonomous robotic exploration using a utility function based on r{\'e}nyi’s general theory of entropy, Autonomous Robots 42~(2) (2018) 235--256.
\newblock \href {https://doi.org/10.1007/s10514-017-9662-9} {\path{doi:10.1007/s10514-017-9662-9}}.

\bibitem{carrillo2012comparison}
H.~Carrillo, I.~Reid, J.~A. Castellanos, On the comparison of uncertainty criteria for active slam, in: 2012 IEEE International Conference on Robotics and Automation, IEEE, 2012, pp. 2080--2087.
\newblock \href {https://doi.org/10.1109/ICRA.2012.6224890} {\path{doi:10.1109/ICRA.2012.6224890}}.

\bibitem{kiefer1974general}
J.~Kiefer, General equivalence theory for optimum designs (approximate theory), The annals of Statistics (1974) 849--879.

\bibitem{carrillo2015monotonicity}
H.~Carrillo, Y.~Latif, M.~L. Rodriguez-Arevalo, J.~Neira, J.~A. Castellanos, On the monotonicity of optimality criteria during exploration in active slam, in: 2015 IEEE International Conference on Robotics and Automation (ICRA), IEEE, 2015, pp. 1476--1483.

\bibitem{rodriguez2018importance}
M.~L. Rodr{\'\i}guez-Ar{\'e}valo, J.~Neira, J.~A. Castellanos, On the importance of uncertainty representation in active slam, IEEE Transactions on Robotics 34~(3) (2018) 829--834.

\bibitem{ahmed2023active}
M.~F. Ahmed, K.~Masood, V.~Fremont, I.~Fantoni, Active slam: A review on last decade, Sensors 23~(19) (2023) 8097.

\bibitem{stachniss2006exploration}
C.~Stachniss, Exploration and mapping with mobile robots, Ph.D. thesis, Verlag nicht ermittelbar (2006).

\bibitem{blanco2008novel}
J.-L. Blanco, J.-A. Fernandez-Madrigal, J.~Gonz{\'a}lez, A novel measure of uncertainty for mobile robot slam with rao—blackwellized particle filters, The International Journal of Robotics Research 27~(1) (2008) 73--89.
\newblock \href {https://doi.org/10.1177/0278364907082610} {\path{doi:10.1177/0278364907082610}}.

\bibitem{amigoni2010information}
F.~Amigoni, V.~Caglioti, An information-based exploration strategy for environment mapping with mobile robots, Robotics and Autonomous Systems 58~(5) (2010) 684--699.
\newblock \href {https://doi.org/10.1016/j.robot.2009.11.005} {\path{doi:10.1016/j.robot.2009.11.005}}.

\bibitem{jadidi2015mutual}
M.~G. Jadidi, J.~V. Miro, G.~Dissanayake, Mutual information-based exploration on continuous occupancy maps, in: 2015 IEEE/RSJ International Conference on Intelligent Robots and Systems (IROS), IEEE, 2015, pp. 6086--6092.
\newblock \href {https://doi.org/10.1109/IROS.2015.7354244} {\path{doi:10.1109/IROS.2015.7354244}}.

\bibitem{ghaffari2018gaussian}
M.~Ghaffari~Jadidi, J.~Valls~Miro, G.~Dissanayake, Gaussian processes autonomous mapping and exploration for range-sensing mobile robots, Autonomous Robots 42 (2018) 273--290.
\newblock \href {https://doi.org/10.1007/s10514-017-9668-3} {\path{doi:10.1007/s10514-017-9668-3}}.

\bibitem{placed2020deep}
J.~A. Placed, J.~A. Castellanos, A deep reinforcement learning approach for active slam, Applied Sciences 10~(23) (2020) 8386.
\newblock \href {https://doi.org/10.3390/app10238386} {\path{doi:10.3390/app10238386}}.

\bibitem{iso1995guide}
{Joint Committee for Guides in Metrology, (JCGM/WG 1)}, Guide to the expression of uncertainty in measurement, Geneva, Switzerland 122 (1995) 16--17.

\bibitem{van2016generalized}
B.~P. Van~Parys, P.~J. Goulart, D.~Kuhn, Generalized gauss inequalities via semidefinite programming, Mathematical Programming 156 (2016) 271--302.
\newblock \href {https://doi.org/10.1007/s10107-015-0878-1} {\path{doi:10.1007/s10107-015-0878-1}}.

\bibitem{thrun2001learning}
S.~Thrun, Learning occupancy grids with forward models, in: Proceedings 2001 IEEE/RSJ International Conference on Intelligent Robots and Systems. Expanding the Societal Role of Robotics in the the Next Millennium (Cat. No. 01CH37180), Vol.~3, IEEE, 2001, pp. 1676--1681.
\newblock \href {https://doi.org/10.1109/IROS.2001.977219} {\path{doi:10.1109/IROS.2001.977219}}.

\bibitem{agha2019confidence}
A.-A. Agha-Mohammadi, E.~Heiden, K.~Hausman, G.~Sukhatme, Confidence-rich grid mapping, The International Journal of Robotics Research 38~(12-13) (2019) 1352--1374.
\newblock \href {https://doi.org/10.1177/0278364919839762} {\path{doi:10.1177/0278364919839762}}.

\bibitem{hershey2007approximating}
J.~R. Hershey, P.~A. Olsen, Approximating the kullback leibler divergence between gaussian mixture models, in: 2007 IEEE International Conference on Acoustics, Speech and Signal Processing-ICASSP'07, Vol.~4, IEEE, 2007, pp. IV--317.
\newblock \href {https://doi.org/10.1109/ICASSP.2007.366913} {\path{doi:10.1109/ICASSP.2007.366913}}.

\bibitem{karaman2011anytime}
S.~Karaman, M.~R. Walter, A.~Perez, E.~Frazzoli, S.~Teller, Anytime motion planning using the rrt, in: 2011 IEEE international conference on robotics and automation, IEEE, 2011, pp. 1478--1483.
\newblock \href {https://doi.org/10.1109/ICRA.2011.5980479} {\path{doi:10.1109/ICRA.2011.5980479}}.

\bibitem{castellanos2004limits}
J.~A. Castellanos, J.~Neira, J.~D. Tard{\'o}s, Limits to the consistency of ekf-based slam, IFAC Proceedings Volumes 37~(8) (2004) 716--721.

\bibitem{sola2010consistency}
J.~Sola, Consistency of the monocular ekf-slam algorithm for three different landmark parametrizations, in: 2010 IEEE International Conference on Robotics and Automation, IEEE, 2010, pp. 3513--3518.

\bibitem{dissanayake2001solution}
M.~G. Dissanayake, P.~Newman, S.~Clark, H.~F. Durrant-Whyte, M.~Csorba, A solution to the simultaneous localization and map building (slam) problem, IEEE Transactions on robotics and automation 17~(3) (2001) 229--241.
\newblock \href {https://doi.org/10.1109/70.938381} {\path{doi:10.1109/70.938381}}.

\end{thebibliography}

\appendix{}
\section{}
\subsection{Uncertainty Mapping proofs}\label{ap:uncertainty_mapping_demo}
Given a measurement with its respective variance-covariance matrix $\mathbf{\Sigma}$, it turns out that:
\begin{align}
    |\mathbf{\Sigma}|=\prod_{i=1}^N \lambda_i,
\end{align}
with $\lambda_i$ the eigenvalues of the matrix $\mathbf{\Sigma}$. These eigenvalues are variances $\sigma_i^2$ in the principal component system. Then:
\begin{align}
    |\mathbf{\Sigma}|=\prod_{i=1}^N  \sigma_i^2 \implies |\mathbf{\Sigma}|^{\frac{1}{2N}}=\left(\prod_{i=1}^N  \sigma_i\right)^{\frac{1}{N}}\triangleq \tilde{\sigma} ,
\end{align}
being $\tilde{\sigma}$ represent the geometric mean of the measurement standard deviations, which quantifies the average uncertainty across all its components.

Considering a vector of independent random variables $\mathbf{X} \in \mathbb{R}^N$ over a $k$-th cell, and a probability density function (pdf) associated with a median vector and a variance-covariance matrix $\mathbf{\Sigma}_k$, there is also an associated $\tilde{\sigma}_k$, then:
\begin{align}
    p_k=P(\mathbf{X}\in A)=\prod_{i=1}^N P\left(|X_i|<\frac{s_i}{2}\right)\geq \prod_{i=1}^N \frac{s_i}{2\sqrt{3}\sigma_i}=\frac{\prod_{i=1}^N s_i}{(2\sqrt{3})^N \tilde\sigma_k^N},
\end{align} 
with $0<s_i\leq 4 \sigma_i /\sqrt{3}$. If $a^N=\frac{\prod_i^N s_i}{(2\sqrt{3})^N}$ is defined, then:
\begin{align}
    p_k\geq \left( \frac{a}{\tilde\sigma_k} \right)^N \implies \tilde\sigma_k \geq \frac{a}{p_k^{1/N}}.
\end{align}
Thus, the $k$-th cell of the uncertainty map $U_k$ is given by:
\begin{align} \label{eq:u_map_apendix}
    U_k=\frac{a}{p_k^{1/N}} \leq \tilde\sigma_k =  |\mathbf{\Sigma}|^{\frac{1}{2N}},
\end{align}
where $a$ is a constant that depends on the interval-length parameter $s_i$. Note that the UM contains lower bounds of the geometric mean of the measurement standard deviations.
\subsection{Numerical example for determining $\beta$}\label{ap:beta}
Assume a Gaussian distribution of a differential robot with a covariance matrix:
\begin{align}
    \mathbf{\Sigma}=\begin{bmatrix}
        \sigma_x^2 &0 & 0 \\
         0 & \sigma_y^2 & 0 \\
         0 & 0 & \sigma_\theta^2 \\
    \end{bmatrix},
\end{align}
where $\sigma_x,\sigma_y$ and $\sigma_\theta$ are the standard deviations of the robot's position and orientation respectively. Then for design requirements the maximun uncertainty for $\sigma_x$ and $\sigma_y$ is $2$m, and for $\sigma_\theta$ is $0.02$rad. On the other hand the hyperrectangle $A$ is defined as $A=s_x \times s_y  \times s_\theta = 0.1m  \times  0.1m  \times 0.002 rad $. With these parameters and using \ref{eq:exploration_map} and \ref{eq:log_odds_prob_p2l} results:
\begin{align}
    &\begin{aligned}[c]
    \beta=1.5863 \times 10^{-5},\  \ell_\beta=-11.051, \ a= \frac{(s_xs_ys_\theta)^{1/3}}{2\sqrt{3}}= 7.8358 \times 10^{-3},  
    \end{aligned}  \\
    &\begin{aligned}[c]
        U_\beta = \frac{a}{\beta^{1/3}}=  0.31186 < |\Sigma_{\text{max}}|^{\frac{1}{6}}=\tilde{\sigma}_{\text{max}}=  0.43089,
    \end{aligned}
\end{align} 
These values are used for update cell algorithms, get SiREn measure and obtain UF when the uncertaity measurement is negligible.

\subsection{Relative entropy: approximations}\label{ap:diver}
In this section, the procedure for obtaining an approximation to calculate the Kullback-Leibler divergence  between two $N-$dimensional Gaussian 
probability densities $f$ and $g$ from the probability map $P_k$ and $Q_k=\beta$ is detailed.

The KL divergence between two probability densities $f$ and $g$ has a closed form \cite{hershey2007approximating} and is given as:
\begin{align}\label{eq:dkl_closed_form}
    D_{\text{KL}}(f||g)=\frac{1}{2}\left( \ln \frac{|\mathbf{\Sigma}_g|}{|\mathbf{\Sigma}_f|}+\text{Tr}(\mathbf{\Sigma}_g^{-1}\mathbf{\Sigma}_f)-n+(\mu_f-\mu_g)^T\mathbf{\Sigma}_g^{-1}(\mu_f-\mu_g) \right),
\end{align}
where $\mathbf{\Sigma}_f$ and $\mathbf{\Sigma}_g$ are the variance-covariance matrices of $f$ and $g$ respectively, $\mu_f$ and $\mu_g$ are the means of $f$ and $g$ respectively 
and $N$ is the dimension of the distributions. Then, taken $\tilde{\sigma}_k $ as geometric mean of the standard deviations of $f$, 
and $\sigma_\text{max}$ as the geometric mean of the standard deviations of $g$, then using (\ref{eq:u_map_apendix}):
\begin{align}
    \text{Tr}(\mathbf{\Sigma}_g^{-1}\mathbf{\Sigma}_f)&=N\left(\frac{\tilde{\sigma}_k}{\sigma_\text{max}}\right)^2\geq N\left(\frac{p_k}{\beta}\right)^{-\frac{2}{N}}  \\
    |\mathbf{\Sigma}_g|=\sigma_{\text{max}}^{2N}&\geq \left(\frac{a^{N}}{\beta}\right)^2  \\
    |\mathbf{\Sigma}_f|=\tilde{\sigma}_k^{2N}&\geq \left(\frac{a^{N}}{p_k}\right)^2.
\end{align} 
Besides, as $\mu_f=\mu_g$, and replacing in (\ref{eq:dkl_closed_form}):
\begin{align}
    D_{\text{KL}}(f||g)&\geq\frac{1}{2}\left( \ln \left(\frac{p_k}{\beta}\right)^2-N+N\left(\frac{p_k}{\beta}\right)^{-\frac{2}{N}} \right) \nonumber\\
    &\geq \ln \frac{p_k}{\beta} -\frac{N}{2} +\frac{N}{2}\left(\frac{\beta}{p_k}\right)^{\frac{2}{N}}.
\end{align}
Note that if $p_k=\beta$ then $D_{\text{KL}}(f\parallel g)=0$, and that  $D_{\text{KL}}(f\parallel g)\geq 0$. From here it is concluded
that, for $p_k$ close enough to $\beta$ this lower bound can be used as an approximation for the Kullback-Leibler divergence.

With similar assumptions, the KL divergence can be obtained as: 
\begin{align}
    D_{\text{KL}}(f||g)=-N\ln \frac{\tilde{\sigma}_k}{\sigma_{\text{max}}}-\frac{N}{2}+\frac{N}{2}\left(\frac{\tilde{\sigma}_k}{\sigma_{\text{max}}}\right)^2.
\end{align}

\subsection{Planning}\label{sec:planning}
As was said in Sec. \ref{sec:uncertainty_frontiers}, an UF is eliminated when $||\nabla U_k||_{2}<T_h$, which happen when the maximum uncertainty of the 
frontiers cells is reduced. 
The uncertainty of a cell only can be reduced when the uncertainty measure propagated by the agent is less than the uncertainty of the cell. For this reason,
the agent's uncertainty must be considered when planning a path to eliminate an UF. The uncertainty of the agent is estimated by the SLAM system, and
depends on the several factors such as the uncertainty of the sensors, the uncertainty of odometry estimations, and so on.

For this reason, a modified version of the well-known RRT* algorithm~\cite{karaman2011anytime} is proposed to plan paths that eliminate an UF. The RRT* algorithm extends an RRT tree by initially growing it as usual. However, before connecting to the nearest node, it searches within a defined radius and selects the connection that minimizes a given cost function. The cost function, $C$, proposed in this work is: 
\begin{align}\label{eq:cost_function}
    C= d+d_{\text{odo}}+\frac{2\tilde{\sigma}_l}{c \tilde{Q}},
\end{align}
where $d$ is the distance between the nodes, $d_{\text{odo}}$ is the distance traveled using only odometry (without references), $c$ is the cell-size of the map and $\tilde{Q}$ a conversion factor
from landmark uncertainty to distance (this value would be the process noise in Kalman filter implementation, ie $|\mathbf{Q}|^{\frac{1}{2N}}=\tilde{Q}$. For this case $\tilde{Q}=0.01$ was chosen). Each landmark has a geometric mean associated $\tilde{\sigma}_l$,
which is calculated as $\tilde{\sigma}_l=|\mathbf{\Sigma}|^{1/2N}$, where  $\mathbf{\Sigma}$  is the variance-covariance matrix of the landmark. 

When the algorithm begins, both $d$ and $\tilde{\sigma}_l$ are initialized to zero. The starting value for $d_{\text{odo}}$ corresponds to the distance traveled using only odometry (without references) by the agent to achieve the current uncertainty. Subsequently, when a node is added to the tree, $d_{\text{odo}}$ is updated to reflect the distance traveled without references from the parent node to the new node. Additionally, $d$ is updated with the distance between the nodes. If a landmark is observed, $d_{\text{odo}}$ is reset to zero, indicating that the uncertainty in the agent's pose estimation is now primarily influenced by the measurement uncertainty. Simultaneously, $\tilde{\sigma}_l$ is updated to match the $\tilde{\sigma}_l$ of the observed landmark. If multiple landmarks are observed, the last $\tilde{\sigma}_l$ is selected. 

\subsection{SLAM system}\label{sec:kf_slam}
Because the main objective of this work is to demonstrate the behavior of UM, the SiREn index, and the UF, an optimal SLAM system was implemented. It is important to note that the focus of this work is not on the SLAM system itself, but rather on ensuring that the covariances of the landmarks and the agent are estimated consistently. It is well known that all variants of nonlinear KF-SLAM estimate overconfident covariances \cite{castellanos2004limits,sola2010consistency}. This issue is crucial to address because decisions are based on the uncertainty represented in the variance-covariance matrix of the map. For these reasons, a linear KF-SLAM system was implemented for positional estimation, and the orientation is considered known, as determined by sensors such as a magnetometer, a gyroscope, among others. 

The robot used in the simulation is a differential robot with a $360^o$ laser scan with a range of $5m$ and a resolution of $0.5^o$. However there are controls systems in a lower level that convert this no holonomic robot in as a holonomic robot for the planning and SLAM system. 

The discrete equations of the KF-SLAM system are: 
\begin{align}
    X[k+1]&=AX[k]+BU[k]+w[k]\\
    Z[k]&=CX[k]+v[k],
\end{align}
where $A$ is an identity matrix, $B=[\mathbf{I}_{2\times 2} ,\quad  \mathbf{0}_{(n-2) \times 2}]$,  $C=[\mathbf{I}_{2\times 2}, \quad  -\mathbf{I}_{2\times 2}]$ is the measure matrix block for each landmark, $U$ is the control input, $w$ and $v$ are the process and measurement noise, respectively. The process and measurement noise is modeled as a Gaussian distribution with zero mean and covariance matrix $\mathbf{Q}=0.01\mathbf{I}$ and $\mathbf{R}=0.01\mathbf{I}$, respectively. 
The covariance matrix of the state is $\mathbf{P}$, and the initial value of the covariance matrix is  $\mathbf{P}_0=0.01\mathbf{I}$. There are not landmaks known at the beginning. 

Finally, no matching errors are considered in the implementation of the SLAM system. This guarantees that the Kalman filter does not diverge. 
For more details about the implementation and convergence of the SLAM system, see~\cite{dissanayake2001solution}.

\end{document}